%% file: gpjournal.tex
\documentclass[10pt]{article}

\usepackage{array,ifthen,amsmath,amssymb,,amsthm,latexsym,color,epsfig,bm,verbatim,graphicx,hyperref,authblk}
\usepackage{natbib}

\input{definitions}

\begin{document}

\title{Subset Selection for Gaussian Markov Random Fields}


\author{Satyaki Mahalanabis\thanks{This research was done while the author was at the University of Rochester.}}
\author{Daniel \v{S}tefankovi\v{c}}
\affil{\{smahalan, stefanko\}@cs.rochester.edu \\ University of Rochester
}


\maketitle

\begin{abstract}
Given a Gaussian Markov random field, we
consider the problem of selecting a subset of variables to observe
which minimizes the total expected squared prediction error of the
unobserved variables. We first show that finding an exact solution is NP-hard
even for a restricted class of Gaussian Markov random fields, called Gaussian
free fields, which arise in semi-supervised learning and computer
vision. We then give a simple greedy approximation  
algorithm for Gaussian free fields on arbitrary graphs. Finally, we give a message passing algorithm for general Gaussian Markov random fields on bounded tree-width graphs. 
\end{abstract}


\section{Introduction}

Given the joint distribution of a set of random variables (in the form of a Markov random field), we
consider the problem of selecting a small subset of these variables to
observe so as to accurately predict the remaining unobserved
variables. We focus here on Gaussian processes~\citep{RW06} on graphs, i.e., Gaussian Markov random fields (Gaussian MRFs). 
 Our aim in this paper is to give a subset selection algorithm which, given a budget for the number of variables that can be observed, minimizes the
expected squared prediction error averaged over all the variables.
We are particularly interested in algorithms with provable guarantees on the prediction error. Our main focus is on Gaussian MRFs on trees and other tree-like graphs, or to be precise, bounded tree-width graphs---such graphs have been widely studied in the context of inference, see, e.g.,~\citet{Sud99}. We also consider a special class of Gaussian MRFs, called Gaussian free fields (or GFFs), which arise, among others, in computer vision, see, e.g.,~\citet{S90}.
We first explain the notation we use and formally state our problem before describing how our work relates to previous research.

\subsection{Notations}

We will use boldface and lowercase to denote a vector, e.g., ${\bf z}$, and use $z_i$ to denote its $i^{th}$ component. We will use uppercase for matrices and random variables (including vectors of random variables). For any $n \times n$ matrix $M$ and subsets $V, V'$ of $\{1, 2 \dots, n\}$, we will use $M[V, V']$ to be the submatrix indexed by rows in $V$ and columns in $V'$. Further, $M[i,:]$ and $M[:,i]$ will denote respectively the $i^{th}$ row and the $i^{th}$ column of $M$. We will say $M$ has support $V \times V'$ if all non-zero entries in $M$ occur in $M[V, V']$. 

For any $k$, $I_k$ will denote the $k \times k$ identity matrix. We will denote the space of $n \times n$ symmetric positive semidefinite matrices (see Chapter~7 of~\citet{HJ85} for a definition) by ${\cal X}_+^{n \times n}$, and the usual ordering on positive semidefinite matrices by $\preceq$ (see Definition~7.7.1 of~\citet{HJ85}). The space of all matrices in ${\cal X}_+^{n \times n}$ with support $V \times V$ will be denoted ${\cal X}_+^{V \times V}$.
We will use ${\cal G}^{n \times n}$ to denote the class of all $n \times n$ symmetric diagonally dominant matrices (see, e.g., Definition 6.1.9 of~\citet{HJ85}) with non-positive off-diagonal entries. Such matrices include, e.g., graph Laplacians.

Given $M \in {\cal X}_+^{n \times n}$, we will use $\lambda_{min}(M)$ and $\lambda_{max}(M)$ to denote respectively the smallest and the largest \emph{non-zero} eigenvalues of $M$, if they exist. Further, if M is of full rank then the condition number of $M$ will mean (see Chapter~5.8 of~\citet{HJ85}) the ratio of the largest to the smallest eigenvalue of $M$.

For random variables ${\bf X} = \big(X_1, X_2, \dots, X_n\big)$ and for any set $V \subseteq \{1,2, \dots, n\}$, we will let ${\bf X}_{V}$ be the coordinates of ${\bf X}$ indexed by $V$, i.e., a vector (of size $|V|$) of the random variables indexed by $V$. If $\rho$ is the joint density function of variables ${\bf X}$, we will use $\ee_{\rho}[\cdot~|~{\bf X}_S]$ and $\var_{\rho}[\cdot~|~{\bf X}_S]$ to denote respectively the conditional expectation and the conditional variance given ${\bf X}_S $ (i.e., given variables in $S$ are observed).  We will drop the subscript $\rho$ when the density is clear from the context.

Some of our algorithms will be fully polynomial time approximation schemes (FPTAS)----see~\citet{Vaz01} for a definition.

\subsection{Definitions and Problem Statement}
\label{sec:gffdef}

A Gaussian MRF on a graph $G = \big(\{1,2, \dots, n\},~E\big)$ is a Gaussian process ${\bf X} = \big(X_1, X_2, \dots, X_n\big)$ with covariance matrix $\Sigma$ (of full rank) such that $E = \big\{\{i, j\}~|~\Lambda_{ij} \neq 0 \big\}$, where $\Lambda \define \Sigma^{-1}$ is the inverse covariance matrix, also called the precision matrix. This means that the joint density of ${\bf X}$, $\rho\big({\bf X}\big)$, is Markov with respect to $G$, i.e., that $\rho\big({\bf X}\big)$ factorizes over the cliques of $G$ (see, e.g.,~\citet{SWW04}). The problem we study assumes the parameters (the means and the covariance matrix) of the Gaussian MRF are known, and hence we assume, w.l.o.g., that ${\bf X}$ is origin-centered.

The following are some well-known facts about Gaussian processes (see, e.g.,~\citet{RW06,Sud99},~also~\citet{KSG08}). Given variables ${\bf X}_S$ to observe, for each $X_i$ the linear predictor ${\bf w}(S,i)^t {\bf X}_S$ which minimizes the expected squared  error, i.e.,
\begin{equation}
\label{e:linpred}
{\bf w}(S,i)~\define~\arg \underset{{\bf w} \in \R^{|S|}}{\min}~\ee\big[\big(X_i - {\bf w}^t {\bf X}_S\big)^2\big],
\end{equation}
is given by ${\bf w}(S,i) = \Sigma[S, S]^{-1}\Sigma[S,i]$, and is an unbiased estimator of $X_i$, i.e., ${\bf w}(S,i)^t{\bf X}_S = \ee\big[ X_i~|~{\bf X}_S\big]$.
The minimum expected squared error for linear prediction for a Gaussian process equals the conditional variance of $X_i$ given ${\bf X}_S$, i.e.,
\begin{equation}
\label{e:varl2}
\begin{split}
\ee\big[\big(X_i - {\bf w}(S,i)^t {\bf X}_S\big)^2\big]~&=~\Sigma[i,i] - \Sigma[i, S]\Sigma[S,S]^{-1}\Sigma[S,i] \\
&=~\ee\bigg[\big(X_i - \ee[X_i | {\bf X}_S] \big)^2~|~{\bf X}_S \bigg]~=~\var\big[ X_i~|~{\bf X}_S\big],
\end{split}
\end{equation}
and is independent of the actual observed values of ${\bf X}_S$. In light of~\eqref{e:varl2}, the average expected squared error for predicting all the  variables,
\begin{equation}
\label{e:gfferr'}
\mathrm{err}\big(S\big)~\define~\frac{1}{n}\sum_{i}~\ee\bigg[\big(X_i - {\bf w}(S,i)^t {\bf X}_S\big)^2\bigg],
\end{equation}
 turns out to be
\begin{equation}
\label{e:gfferr}
\mathrm{err}\big(S\big)~=~\frac{1}{n}\sum_{i}~\var\big[ X_i~|~{\bf X}_S\big]~=~\frac{1}{n}\sum_{i \notin S}\var\big[ X_i~|~{\bf X}_S\big],
\end{equation}
and can also be expressed as
\begin{equation}
\label{e:erreigen}
\begin{split}
\mathrm{err}\big(S\big)~&=~\frac{1}{n}\TR\bigg(~\Sigma[\overline{S}, \overline{S}] - \Sigma[\overline{S}, S]\big(\Sigma[S, S] \big)^{-1}\Sigma[S, \overline{S}] \bigg) \\
~&=~\frac{1}{n}\TR\big(\Lambda[\overline{S}, \overline{S}]^{-1}\big)
\end{split}
\end{equation}

Our error function $\mathrm{err}(S)$, and hence $\var\big[ X_i ~|~ {\bf X}_S\big]$, is clearly monotone decreasing in $S$. In fact we can show that the expected conditional variance $\ee\big[\var\big[ X_i ~|~ {\bf X}_S\big] \big]$, for any set of random variables $X_1, X_2, \dots, X_n$ (not necessarily Gaussian) is monotone decreasing.
\BLEM
\label{l:monodec}
For any subsets $S \subseteq T$ and for any $X_i$, $\ee\big[ \var\big[ X_i ~|~ {\bf X}_S\big] \big] \geq \ee\big[\var\big[X_i~|~{\bf X}_T\big] \big]$.
\ELEM

\BPRF[of Lemma~\ref{l:monodec}:] 
We will use the identity that for any sets $S$ and $T$,
\begin{equation}
\label{e:monotone}
\ee\big[\var\big[ X_i~|~{\bf X}_{S \cup T} \big] ~|~ {\bf X}_S\big] = \var\big[X_i~|~{\bf X}_S\big] - \var\big[ \ee\big[ X_i~|~{\bf X}_{S \cup T} \big]~|~{\bf X}_S\big].
\end{equation}
Now the lemma follows from the fact that $S \subseteq T$ and taking expectation of both sides of~\eqref{e:monotone} w.r.t. ${\bf X}_S$.
\EPRF

We formally state the problem of finding a variable selection strategy below.
\BQUE
\label{que:gffselect}
Is there an algorithm which given a budget $b$ finds a set $S^*$ of size $|S^*| \leq b$ such that 
\begin{equation}
\label{e:gffopt}
\mathrm{err}\big(S^*\big)~=~\min_{|S| \leq b}~\mathrm{err}\big(S\big).
\end{equation}
\EQUE

\BQUE 
\label{que:gffselectcov}
The cover version of Question~\ref{que:gffselect} is whether an algorithm exists which, given $\alpha > 0$, finds the smallest $S^*$ such that 
\begin{equation}
\label{e:gffcover}
|S^*| = \min~\big\{ |S|~|~\mathrm{err}\big(S\big) \leq \alpha\big\}.
\end{equation}
\EQUE

A Gaussian free field or GFF (see, e.g., Chapter~2.7 of~\citet{LP11}) is a special case of a Gaussian MRF on a connected graph $G$, where each edge $\{i,j\} \in E$ is associated with a finite weight $r_{ij} = r_{ji} > 0$. We assume that for each $\{i, j\} \notin E$, $r_{ij}  = r_{ji} = +\infty$. Fix a node, say $1$, and assume that $X_1 = 0$. Then the density of a GFF is given by
\begin{equation}
\label{e:gff}
\rho(X_1, X_2, \dots, X_n)~\propto~\exp\bigg( \sum_{\{i, j\} \in E} -\frac{(X_i - X_j)^2}{2r_{ij}} \bigg). 
\end{equation}

We set one variable, namely $X_1$, to $0$ so that the density $\rho$ in~\eqref{e:gff} is well defined. This corresponds to always selecting the variable $X_1$ for observation. Consider the Laplacian $\Lambda$ 
\begin{equation}
\label{e:gffLaplacian}
\Lambda[i,j]~\define~\left\{ \begin{array}{cc}
\sum_{t \neq i}~\big(1~/~r_{it}\big)~&~\textrm{if}~i = j \\
-1~/~r_{ij}~&~\textrm{otherwise},
\end{array} \right. 
\end{equation}
Although a GFF really defines a distribution on ${\bf X}_{\{2, \dots, n\}}$ with precision $\Lambda[\{2, \dots, n\}, \{2,\dots, n\}]$, we will work with the matrix $\Lambda$ so that we can treat $X_1$ symmetrically with the other variables. Note that any Gaussian distribution whose precision matrix is strictly diagonally dominant with non-positive off-diagonal entries can be thought of as the marginal distribution (over $n$ variables) in a GFF with $n+1$ variables.

\subsection{Summary of Contributions}
We list the main results in this paper in the order they are presented:

\begin{itemize}
\item Finding an exact solution to Question~\ref{que:gffselect} is NP-hard even for GFFs (Section~\ref{sec:supermodular}, Theorem~\ref{t:gffnph}).

\item The average expected squared error function $\mathrm{err}$ is supermodular for GFFs, thereby giving greedy approximation algorithms for  Questions~\ref{que:gffselect} and~\ref{que:gffselectcov} (Section~\ref{sec:supermodular}, Theorems~\ref{t:gffbudget} and~\ref{t:gffapprox}).


\item There is a FPTAS for GFFs on bounded tree-width graphs based on message passing (i.e., dynamic programming). While it is not difficult to formulate a dynamic programming algorithm for Gaussian MRFs on trees (which are equivalent to a GFFs on trees), extending it to bounded tree-width graphs is non-trivial and requires a more intricate analysis of the error (Section~\ref{sec:balancedtree}, Theorem~\ref{t:gfftwmain}).

\item There is a similar FPTAS based on dynamic programming for general Gaussian MRFs on bounded tree-width graphs, whose running time however scales as a polynomial in the condition number of the input precision matrix (Section~\ref{sec:GPtreewidthapprox}, Theorem~\ref{t:gpmain}). 

\end{itemize} 

\subsection{Related Work}
\label{sec:gffprev}

We first compare our error function (i.e., the average expected squared error) with those used earlier in the context of subset selection for Gaussian Processes. Prediction for Gaussian processes is popularly known as kriging in spatial statistics (see, e.g.,~\citet{RW06}), and the squared prediction error for an unobserved variable is referred to as its kriging variance. The problem we try to solve can be thought of as minimizing the average kriging variance (i.e., average expected squared error) over all the unobserved variables. A closely related work is~\citet{KMGG07}, who consider the problem of minimizing the maximum kriging variance of an unobserved variable rather than the average. Also, there is extensive literature exploring subset selection for Gaussian processes using different criteria like entropy and mutual information between observed and unobserved variables, see, e.g.,~\citet{KSG08, KG11}. 

Subset selection problems similar to ours arise in a number of other areas. For instance, our problem is quite similar to the widely studied problem of subset selection for regression in statistics, see, e.g.,~\citet{M02} for an overview. Our objective differs from subset selection in regression in that we aim to minimize the total prediction error of all the unobserved variables rather than that of a single variable. Recently~\citet{DK08, DK11} have analyzed and provided provable guarantees for several well-known greedy algorithms for subset selection,  such as forward selection. 

 Our goal of minimizing the average expected squared error of unobserved variables is also equivalent to minimizing the trace of the inverse of a principal submatrix of the inverse covariance matrix. A similar trace minimization problem arises in Bayesian and transductive experimental design, see, e.g.,~\citet{CV95, YBT06}. For a linear model with a Gaussian prior over the unknown parameters, the Bayesian A-optimality criterion~\citep{CV95} reduces to minimizing the trace of the conditional covariance matrix of the parameters given the selected experiments (observations).  However, our objective differs from A-optimality in that we want to minimize the prediction error of all the unobserved variables given the observed ones. In contrast,  A-optimality requires minimizing the error of only a given subset of unobserved variables (i.e., the parameters being estimated), and moreover none of the parameter values can be observed.
 
The budget version of our problem, Question~\ref{que:gffselect}, can alternatively be formulated (using expression~\eqref{e:erreigen} for the error) as finding a low rank approximation of the positive semidefinite covariance matrix $\Sigma$, or to be precise, a rank $b$ Nystr\"{o}m approximation (see, e.g.,~\citep{WS01}) of $\Sigma$ which minimizes the trace norm error. Nystr\"{o}m approximation has applications in Gaussian process regression, kernel machines and dimension reduction among others (see, e.g.,~\citet{SS00,WS01}). However, to the best our knowledge, we are the first to focus on subset selection for bounded tree-width Gaussian MRFs (including GFFs)---we exploit the sparse structure of the precision matrices for such MRFs to give a dynamic programming based algorithm. Subset selection strategies for Nystr\"{o}m approximation based on greedy heuristics (see, e.g.,~\citet{SS00}) and on random sampling (see, e.g.,~\citet{DM05,KMT09}) have been studied before. In contrast the subset selection strategies we give in this paper are deterministic, have provable error bounds, and moreover have multiplicative rather than additive approximation guarantees for the error.

We next discuss our motivation for considering bounded tree-width Gaussian MRFs and Gaussian free fields. Our motivation for studying bounded tree-width graphs comes from the fact that the exact solution of many problems (such as finding a maximum independent set and inference in graphical models, see, e.g.,~\citet{Bod97,KF09}), which are infeasible in general, become tractable for bounded tree-width graphs.  There is extensive literature on inference algorithms for graphical models, especially Gaussian MRFs---we refer to Chapter~2 of~\citet{Sud99} for a survey. We point put that though inference for Gaussian MRFs can be performed in polynomial time (since it involves mainly a matrix inversion) for any graph, our problem of subset selection is much harder. In fact, as we show later, our problem is NP-hard even for the restricted case of Gaussian free fields. Also note that our problem is harder than, e.g., maximum independent set, in the sense that our problem does not fit into the framework of monadic second order logic on graphs~\citep{Cou90, Bod97}.

A Gaussian free field (or GFF) is a special case of a Gaussian MRF, and can be thought of as ``continuous analog" of a ferromagnetic Ising Model. A GFF  corresponds to the inverse covariance matrix being a graph Laplacian, and is widely used in semi-supervised learning, (see, e.g.,~\citet{ZLG03, BN04}) as well as in computer vision, e.g.,~\citep{S90}. In~\citet{ZLG03WS} a GFF is used to model the distribution of a discrete binary MRF, i.e., an Ising Model, given a set of observed labels. Specifically, the expected value of an unobserved node in the GFF is used as an approximation for the node's expected (binary) label value. A simple adaptive selection strategy is then given which queries the node with the most uncertain label. Our aim here, unlike that of~\citet{ZLG03WS}, is to (non-adaptively) select subsets for prediction in a GFF rather than use a GFF  for adaptive selection in Ising Models.

\section{Gaussian Free Field}
\label{sec:supermodular}

We prove that finding an exact solution to Question~\ref{que:gffselect} is not feasible even for GFFs. Note that the related problem of subset selection for regression is NP-hard~\citep{N95} and is also hard to approximate within a constant factor when the subset size is $\Theta(\log n)$~\citep{DK08}. However our infeasibility proof is for the special case of GFF, and hence does not follow from the hardness proofs for subset selection. We need the following characterization of our problem for regular graphs.
 
\BLEM
\label{l:gffcov}
Let graph $G$ be $d$-regular and let each edge $\{i,j\}$ have weight $r_{ij} = 1$. Then for any set $S$ of nodes, $\mathrm{err}\big(S\big) \geq \big(1 - |S| / n\big)/d$ where the equality holds iff $\overline{S}$ is an independent set in $G$.
\ELEM

\BPRF[of Lemma~\ref{l:gffcov}:]
Since $G$ is $d$-regular and each edge has weight $1$, each diagonal entry of $\Lambda$ is $d$. Hence for any set $S$,
\begin{equation}
\label{e:gffcov1}
\TR\big(\Lambda[\overline{S}, \overline{S}]\big)~=~(n - |S|)d~=\sum_{i=1}^{n-|S|}~\lambda_i\big(\Lambda[\overline{S}, \overline{S}]\big),
\end{equation} 
where $\lambda_i\big(\Lambda[\overline{S}, \overline{S}]\big)$ are the eigenvalues of $\Lambda[\overline{S}, \overline{S}]$ in any order.
Using expression~\eqref{e:erreigen} for the error, we have by~\eqref{e:gffcov1} that
\begin{equation}
\label{e:gffcov3}
\begin{split}
\mathrm{err}(S)~=~\frac{1}{n}\TR\big(\Lambda[\overline{S}, \overline{S}]^{-1}\big)~&=~\frac{1}{n}\sum_{i=1}^{n - |S|}~\frac{1}{\lambda_i\big(\Lambda[\overline{S}, \overline{S}]\big)} \\
~&\geq~\frac{1}{n}\frac{(n-|S|)^2}{\sum_{i=1}^{n-|S|}~\lambda_i\big(\Lambda[\overline{S}, \overline{S}]\big)}~=~(1 - |S|/n) / d,
\end{split}
\end{equation}
where in the third step we used the fact that arithmetic mean is greater than harmonic mean, which are equal only if each of the elements (i.e., each eigenvalue of $\Lambda[\overline{S}, \overline{S}]$) are equal, i.e.,
where equality holds iff 
\begin{equation}
\label{e:gffcov2}
\lambda_1\big(\Lambda[\overline{S}, \overline{S}]\big) = \lambda_2\big(\Lambda[\overline{S}, \overline{S}]\big) = \dots = \lambda_{n-|S|}\big(\Lambda[\overline{S}, \overline{S}]\big) = d. 
\end{equation}
However~\eqref{e:gffcov2} holds iff $\Lambda[\overline{S}, \overline{S}]$ is a diagonal matrix, i.e., $\overline{S}$ is an independent set. The Lemma now follows.
\EPRF
 
\BTHM
\label{t:gffnph}
The budget version~\eqref{e:gffopt} of Question~\ref{que:gffselect} is NP-hard.
\ETHM

\BPRF[of Theorem~\ref{t:gffnph}:] 
The problem of finding, for any $n,k,d$ and any $d$-regular graph on $n$ nodes, whether an independent set of size $k$ exists, is known to be NP-complete, see, e.g., problem [GT20] in~\citet{GJ00}. In fact, finding independent sets is NP-complete even for bounded-degree planar graphs.  It follows from Lemma~\ref{l:gffcov} that  an independent set of size $n-b$ exists iff for $S^*$ as defined  in~\eqref{e:gffopt} and for budget $|S^*| \leq b$, we have $\mathrm{err}\big(S^*\big) = (1 - b/n)/d$.
\EPRF
 
We now give an approximation algorithm for Question~\ref{que:gffselect} using supermodularity of the average expected  squared error. Since we directly consider the variance instead of variance reduction (i.e, the $R^2$-statistic, see, e.g.,~\citet{DK08}), it is more convenient to use the notion of supermodularity rather than submodularity.
\BDEF 
A function $f : 2^{\{1, 2, \dots, n\}} \to \R$ is supermodular if the following ``diminishing returns" condition holds:
\begin{equation}
\label{e:supermoddef}
(\forall A)~(\forall x,y) \quad f\big(A\big) - f\big(A \cup \{x\} \big) \geq f\big(A \cup \{y\}\big) - f\big(A \cup \{x,y\}\big).
\end{equation}
\EDEF

Given oracle access to a non-negative supermodular function $f : 2^{\{1, 2, \dots, n\}} \to \R^+$ such that $f(\{1,2, \dots, n\}) = 0$ and $(\forall A,B)~A \subseteq B \Rightarrow  f(A) \geq f(B)$ (i.e., $f$ is monotone non-increasing), the supermodular minimization problem is to compute $S_b^*~\define~\arg \underset{|S| \leq b}{\min}~f(S)$, while the supermodular cover problem is to compute $
S_{\alpha}^*~\define~\arg \underset{f(S) \leq \alpha}{\min}~|S|$. We refer to~\citet{NWF78,Wol82} for the well-known greedy algorithm for both constrained minimization and cover problems for supermodular functions. \\

Our main result in this section is a proof that the error function $\mathrm{err}$ is supermodular, for which we need the following well-known connection between electrical networks and GFFs.

\BLEM (see, e.g., Lemma~2.15 of~\citet{DLP11})
\label{l:varres}
Consider the GFF given by~\eqref{e:gff}, any non-empty set $S$ and any $i \notin S$.  Then 
$$\var\big[ X_i~|~{\bf X}_S\big] = R_{\mathrm{eff}}\big(i,S\big),$$
where $R_{\mathrm{eff}}\big(i,S\big)$ is the effective resistance between $i$ and $S$ in $G$ with each $r_{ij}~(= r_{ji})$ being interpreted as the resistance of edge between $i,j$.
\ELEM

\BLEM
\label{l:gffsub}
For a GFF the error function $\mathrm{err}\big(S\big)$, or equivalently the function $\TR\big(\Lambda[\overline{S}, \overline{S}]^{-1}\big)$ (using~\eqref{e:erreigen}), is non-increasing and supermodular in $S$.
\ELEM
We defer the proof to the end of this section. The following example illustrates that Lemma~\ref{l:gffsub} is not true for Gaussian processes in general.
\BEX
For the multivariate Gaussian with covariance matrix
$$ \Sigma = \left[ \begin{array}{cccc}  
    0.4435  &  0.1092  & -0.0905 & -0.0527 \\
    0.1092  &  0.3041  & 0.0256  & 0.0227 \\
   -0.0905  &  0.0256  & 0.1273  & -0.1444 \\
   -0.0527  &  0.0227  & -0.1444 & 0.3752 
\end{array} \right],
$$
$\mathrm{err}\big(\{1\}\big) = 0.1887,  \mathrm{err}\big(\{1,2\}\big) = 0.1162, \mathrm{err}\big(\{1, 3\}\big) = 0.1009$ and $\mathrm{err}\big(\{1,2,3\}\big) =  0.0263$ which violates supermodularity\footnote{all figures precise up to $\pm 10^{-4}$}.
\EEX

Submodularity has been used previously (see, e.g.,~\citep{KSG08}) for a similar problem, namely sensor placement, where the objective (unlike ours) is usually to maximize the mutual information between observed and unobserved nodes. We also point out that in~\citet{DK08}, it is shown that the absence of suppressor variables is a necessary and sufficient condition for the error to be supermodular (see Theorem~8.1 of~\citet{DK08} for details). Our proof of supermodularity does not use their result---instead we use the analogy with electrical networks. Note that Lemma~\ref{l:gffsub} implies that a GFF does not have any suppressor variables in the sense of~\citet{DK08}. 

In~\citet{DK11}, a sufficient condition for supermodularity of $\var\big[ X_i ~|~ {\bf X}_S\big]$ for a Gaussian process (see Definition~2.3 and Lemma~2.4) is given in terms of eigenvalues of covariance matrix. However as Example~\ref{exm:cegdk11} shows, their result does not apply to Lemma~\ref{l:gffsub}.
\BEX
\label{exm:cegdk11}
Consider a GFF on the complete graph on $5$ nodes with weights $(\forall i \neq j)~r_{ij} = 5/2$ and with $X_1 = 0$. Then $(\forall i > 1)~\var\big[X_i\big] = 1$, but the covariance matrix between $X_2, X_3, X_4$ has a minimum eigenvalue $0.5$. Hence Lemma~2.4 of~\citet{DK11} does not imply supermodularity of $\var\big[ X_i ~|~ {\bf X}_S\big]$ for this GFF.
\EEX

Given Lemma~\ref{l:gffsub}, an answer to the budget version follows immediately from the greedy algorithm of~\citet{NWF78}.
\BTHM
\label{t:gffbudget}
For the budget version~\eqref{e:gffopt} of Question~\ref{que:gffselect}, there is a greedy algorithm which for any budget $b$, outputs a set $S_b$ such that $\mathrm{err}\big(S_b\big)~\leq \frac{1}{(1 - 1 / \e)}~\min_{|S| \leq b}~\mathrm{err}\big(S\big)$.
\ETHM

The proof for cover version (Question~~\ref{que:gffselectcov}), Theorem~\ref{t:gffapprox}, requires Lemma~\ref{l:wolsey}.

\BLEM
\label{l:wolsey}
(see Theorem~1 of~\citet{Wol82}) There exists an algorithm which for any non-negative supermodular non-increasing function $f$ and any $\alpha > 0$ computes set $S_{\alpha}$ such that  $f(S_{\alpha}) \leq \alpha$ and $|S_{\alpha}| \leq (1 + \kappa) | S_{\alpha}^*|$, where 
\begin{equation}
\label{e:wolsey}
\kappa = \ln\bigg( \max_{x,S}\bigg\{ \frac{f(\emptyset) - f(\{x\})}{f(S) - f(S \cup \{x\})}~\big|~f(S) > f(S \cup \{x\}) \bigg\}\bigg).
\end{equation}
\ELEM

\BTHM
\label{t:gffapprox}
There exists an algorithm which given any GFF and any $\alpha > 0$, outputs a set $S_{\alpha}$ such that  $\mathrm{err}(S_{\alpha}) \leq \alpha$ and 
\begin{equation}
\label{e:gffapprox}
|S_{\alpha}| \leq  \bigg(1 + \ln\bigg((n-1)^2 \frac{R}{r} \bigg)\bigg) \cdot 
 \min_{\mathrm{err}(S) \leq \alpha}~|S|,
\end{equation}
where $r = \min_{i \neq j}~r_{ij}$ and $R = \max_{\{i,j\} \in E}~r_{ij}$.
\ETHM

\BPRF[of Theorem~\ref{t:gffapprox}:] 
Since by Lemma~\ref{l:gffsub} the function $\mathrm{err}$ is non-increasing and supermodular, we can apply Lemma~\ref{l:wolsey}. It only remains to be prove an appropriate upper bound on the r.h.s. of~\eqref{e:wolsey}. We will show that for any~$x,S$ s.t. $x \notin S$ (and keeping in mind that we assume $1 \in S$),
\begin{equation}
\label{e:gffapprox3}
\frac{\mathrm{err}(\{1\}) - \mathrm{err}(\{1,x\})}{\mathrm{err}(S) - \mathrm{err}(S \cup \{x\})} \leq \frac{(n-1)^2R}{r}.
\end{equation}
We will use the fact that by Lemma~\ref{l:varres} and~\eqref{e:gfferr}, 
$$\mathrm{err}\big(S\big) = \frac{1}{n}\sum_{i \notin S}~R_{\mathrm{eff}}\big(i,S\big).$$
We can  bound 
\begin{equation}
\label{e:gffapprox1}
\begin{split}
\mathrm{err}(S) - \mathrm{err}(S \cup \{x\}) &~\geq~\mathrm{err}(\{1,2, \dots, n\} \setminus \{x\}) - \mathrm{err}(\{1,2,\dots,n\}) \\
& \qquad \qquad \qquad (\textrm{since}~\mathrm{err}~\textrm{is supermodular}) \\ 
&~=~\mathrm{err}(\{1,2, \dots, n\} \setminus \{x\})\\
&~=~R_{\mathrm{eff}}\big(x, \{1,2, \dots, n\} \setminus \{x\}\big)~\geq~r / (n-1),
\end{split}
\end{equation}
since there can be, in the worst case, $n-1$ edges of resistance $r$ between $x$ and the rest of the nodes.
Also we have 
\begin{equation}
\label{e:gffapprox2}
\mathrm{err}(\{1\}) - \mathrm{err}(\{1,x\})~\leq~\mathrm{err}(\{1\})~\leq~\max_{y}~R_{\mathrm{eff}}\big(y,\{1\}\big)~\leq~(n-1)R.
\end{equation}
From~\eqref{e:gffapprox1} and~\eqref{e:gffapprox2} we get~\eqref{e:gffapprox3}, which gives the desired upper bound on~\eqref{e:wolsey}.
\EPRF

Hence the subset chosen by our algorithm is only $O(\ln n)$ times the optimal. For proving Lemma~\ref{l:gffsub} we need Lemmas~\ref{l:thomson} and~\ref{l:flow}.
\BLEM
\label{l:thomson}
(Thomson's Principle, see, e.g., Chapter 2.4 of~\citet{LP11})~
For any unit flow $f$ in $G$ from a set of nodes $S$ to any node $t \notin S$, define 
$$\mathcal{E}(f) \define \frac{1}{2}~\sum_{i \neq j}~f(i,j)^2 r_{ij}.$$
Then
$$\var\big[ X_t~|~{\bf X}_S\big] = \min_{f}~\mathcal{E}(f).$$ 
\ELEM
\BPRF[of Lemma~\ref{l:thomson}:] 
From Thomson's Principle  we have that $R_{\mathrm{eff}}\big(t,S\big) = \min_{f}~\mathcal{E}(f)$ (see, e.g., Chapter 2.4 of~\citet{LP11}, also Lemma~2.11 of~\citet{DLP11}).  Our claim now follows from Lemma~\ref{l:varres}.
\EPRF

The next lemma states that given 2 unit flows in a network from 2 sources $S$ and $T$ to a common sink, one can always construct 2 other unit flows from sets $S \cup T$ and $S \cap T$ to the sink.

\BLEM
\label{l:flow}
Consider any set $S$ and any distinct $x,y,t \notin S$.  Then for any 2 unit flows $f_1, f_2$ in G respectively from $S, S \cup \{x,y\}$  to $t$, there exist 2 corresponding unit flows $g_1, g_2$ respectively from $S \cup \{x\}, S \cup \{y\}$  to $t$ such that for each $i \neq j$, 
\begin{equation} 
\label{e:gffsub2}
g_1(i,j) + g_2(i,j) = f_1(i,j) + f_2(i,j)~\textrm{and},
\end{equation} 
\begin{equation} 
\label{e:gffsub3}
\begin{split}
\min\{f_1(i,j), f_2(i,j)\} &\leq g_1(i,j) \leq \max\{f_1(i,j), f_2(i,j)\}, \\
\min\{f_1(i,j), f_2(i,j)\} &\leq g_2(i,j) \leq \max\{f_1(i,j), f_2(i,j)\}.
\end{split}
\end{equation}
\ELEM
We defer the proof of Lemma~\ref{l:flow} to the end of this section. We now have all the ingredients for proving our main result.

\BPRF[of Lemma~\ref{l:gffsub}:] 
Lemma~\ref{l:monodec} implies that $\mathrm{err}(S)$ is non-increasing.

We will prove that for any $t$, $\var\big[ X_t~|~{\bf X}_S\big]$ is supermodular, which by~\eqref{e:supermoddef}, requires us to show that for any set $S$ and any 2 nodes $x,y \notin S$, 
\begin{equation}
\label{e:gffsub1}
\var\big[X_t~|~{\bf X}_S\big] + \var\big[ X_t~|~{\bf X}_{S \cup \{x,y\} }\big] 
\geq~\var\big[ X_t~|~{\bf X}_{S \cup \{x\}} \big] + \var\big[ X_t~|~{\bf X}_{S \cup \{y\}} \big].
\end{equation}
The supermodularity of $\mathrm{err}(S)$ follows from~\eqref{e:gffsub1} since $\mathrm{err}(S) =  \frac{1}{n}\sum_{i}~\var\big[ X_i~|~{\bf X}_S\big]$. If $t \in S \cup \{x,y\}$ then~\eqref{e:gffsub1} follows easily from Lemma~\ref{l:monodec}. 

Now assume $t \notin S \cup \{x,y\}$. By Lemma~\ref{l:flow}, for any 2 unit flows $f_1, f_2$ in G from $S, S \cup \{x,y\}$ respectively to $t$,  there exist unit flows $g_1, g_2$ from $S \cup \{x\}, S \cup \{y\}$ respectively to $t$ satisfying~\eqref{e:gffsub2} and~\eqref{e:gffsub3}. It follows from~\eqref{e:gffsub2},~\eqref{e:gffsub3} that for each $i \neq j$
$$f_1^2(i,j) + f_2^2(i,j)~\geq~g_1^2(i,j) + g_2^2(i,j),$$ 
which implies
\begin{equation}
\label{e:gffsub4}
\begin{split}
\mathcal{E}(f_1) + \mathcal{E}(f_2) &= \frac{1}{2} \sum_{i \neq j} \big( f_1^2(i,j) + f_2^2(i,j) \big) r_{ij} \\ 
& \geq \frac{1}{2} \sum_{i \neq j} \big( g_1^2(i,j) + g_2^2(i,j) \big) r_{ij} \\ 
& = \mathcal{E}(g_1) + \mathcal{E}(g_2).
\end{split}
\end{equation}
We can now combine Lemma~\ref{l:thomson} with~\eqref{e:gffsub4} to obtain~\eqref{e:gffsub1}. \\
\EPRF

Finally, we finish off with the proof of our flow composition lemma, Lemma~\ref{l:flow}.

\BPRF[of Lemma~\ref{l:flow}:] 
To prove that $g_1$ exists, we note that $g_1$ is a feasible solution to the linear program defined in~\eqref{e:gffsub5}. Existence of $g_2$ follows from that of $g_1$ as $(\forall i \neq j)~g_2(i,j) \define f_1(i,j) + f_2(i,j) - g_1(i,j)$. 
\begin{equation}
\label{e:gffsub5}
\begin{split}
\textrm{maximize}& \quad 0, \quad \textrm{subject to} \\
\textrm{antisymmetry:}& \quad (\forall i < j)~g_1(i,j) + g_1(j,i) = 0, \\
\textrm{flow conservation:}& \quad (\forall i \notin S \cup \{x, t\})~\sum_j~g_1(i,j) = 0, \\
\textrm{unit flow:}& \quad \sum_j~g_1(j, t) = 1, \\
\textrm{value of}~f_2~\textrm{at}~x\textrm{:}& \quad \sum_j~g_1(x,j) = Q \define \sum_{j}~f_2(x,j), \\
\textrm{capacity constraint:}& \\
(\forall i \neq j)~g_1(i,j) & \leq C_{ij} \define \max\{f_1(i,j), f_2(i,j)\}.
\end{split}
\end{equation}

Note that capacity constraint, antisymmetry and~\eqref{e:gffsub2} together imply~\eqref{e:gffsub3}.  Consider the dual of~\eqref{e:gffsub5}, with dual variables $\{d(\{i,j\})\}_{i \neq j}$ (antisymmetry), $\{a(i)\}_{i \notin S \cup \{x, t\}}$ (flow conservation), $a(t)$ (unit flow), $a(x)$ (value of $f_2$ at $x$) and $\{ b(i,j) \}_{i \neq j}$ (capacity constraint).
There exists of feasible solution to~\eqref{e:gffsub5} iff the dual objective~\eqref{e:gffsub6} is non-negative.
\begin{equation}
\label{e:gffsub6}
\begin{split}
\textrm{minimize}&~ a(t) - Q a(x) + \sum_{i \neq j}~C_{ij}b(i,j), \quad \textrm{subject to}\\
&~(\forall i \neq j)~b(i,j) + d(\{i,j\}) +  a(i) =  0, \\
&~(\forall i \in S)~a(i) = 0, \\
&~(\forall i\neq j)~b(i,j) \geq 0.
\end{split}
\end{equation}
Eliminating $\{ d(\{i,j\}) \}_{i \neq j}$ from~\eqref{e:gffsub6} yields
\begin{equation}
\label{e:gffsub7}
\begin{split}
\textrm{minimize}&~ a(t) - Q a(x) + \sum_{i \neq j}~C_{ij}b(i,j), \quad \textrm{subject to}\\
&~(\forall i < j)~b(i,j) + a(i) = b(j,i) + a(j), \\
&~(\forall i \in S)~a(i) = 0, \\
&~(\forall i\neq j)~b(i,j) \geq 0.
\end{split}
\end{equation}
Further eliminating $\{b(i,j)\}_{i \neq j}$ from~\eqref{e:gffsub7} gives~\eqref{e:gffsub8}.
\begin{equation}
\label{e:gffsub8}
\begin{split}
\textrm{minimize}&~ a(t) - Q a(x) + \sum_{a(j) \geq a(i)}C_{ji}(a(j) - a(i)) \\ 
\textrm{subject to}~&~(\forall i \in S)~a(i) = 0~\textrm{and}~(\forall i)~a(i) \in [-1,1],
\end{split}
\end{equation}
where bounding each $a(i)$ to be in $[-1, 1]$ does not affect the sign of the objective~\eqref{e:gffsub8}. 
In fact, by the same argument as that in the proof of the max-flow min-cut theorem (see, e.g., Chapter~12.2 of~\citet{Vaz01})~\eqref{e:gffsub8} is equivalent to the following integer program
\begin{equation}
\label{e:gffsub8'}
\begin{split}
\textrm{minimize}&~ a(t) - Q a(x) + \sum_{a(j) \geq a(i)}C_{ji}(a(j) - a(i)) \\ 
\textrm{subject to}~&~(\forall i \in S)~a(i) = 0~\textrm{and}~(\forall i)~a(i) \in \{-1,0,1\}.
\end{split}
\end{equation}
Applying the shift $(\forall i)~\textrm{s.t.}~a(i) \neq 0,~a(i) \mapsto a(i)+z$, where $~z \in [-1, 1]$ to~\eqref{e:gffsub8'}, gives us an objective which is linear in $z$ and whose minimum is achieved at either $z = -1$ or $z = 1$. Hence the minimum in~\eqref{e:gffsub8'} can not be less than that of~\eqref{e:gffsub8''}. 
\begin{equation}
\label{e:gffsub8''}
\begin{split}
\textrm{minimize}&~ a(t) - Q a(x) + \sum_{a(j) \geq a(i)}C_{ji}(a(j) - a(i)) \\ 
\textrm{subject to}&~(\forall i \in S)~a(i) = 0,~\textrm{and}~\textrm{either}~(\forall i)~a(i) \in \{-2,0\}~\textrm{or}~(\forall i)~a(i) \in \{0,2\}.
\end{split}
\end{equation}
Further the objective~\eqref{e:gffsub8''} is non-negative iff~\eqref{e:gffsub9} is non-negative where we have just scaled each $a(i)$ by $1/2$.
\begin{equation}
\label{e:gffsub9}
\begin{split}
\textrm{minimize}&~ a(t) - Q a(x) + \sum_{a(j) \geq a(i)}C_{ji}(a(j) - a(i)) \\ 
\textrm{subject to}&~(\forall i \in S)~a(i) = 0,~\textrm{and}~\textrm{either}~(\forall i)~a(i) \in \{-1,0\}~\textrm{or}~(\forall i)~a(i) \in \{0,1\},
\end{split}
\end{equation} 
We first assume that $(\forall i)~a(i) \in \{-1,0\}$ and show that the minimum in~\eqref{e:gffsub9} is non-negative.  
Any assignment $(\forall i)~a(i) \in \{-1,0\}$ defines a cut in G between nodes $\{i ~|~a(i) = 0\}$ (including $S$) and nodes $\{i ~|~a(i) = -1\}$. If $a(t) = 0$, 
\begin{equation*}
a(t) - Q a(x) + \sum_{a(j) \geq a(i)}C_{ji}(a(j) - a(i))~\geq~\sum_{a(j) \geq a(i)}C_{ji}(a(j) - a(i))~\geq~\sum_{\substack{a(j) = 0 \\ a(i) = -1}}f_1(j,i) = 0,
\end{equation*}
since $(\forall i \neq j)~C_{ji} \geq f_1(j,i)$ and $f_1$ is a flow from $S$ to $t$ which are all on the same side of the cut.
On the other hand, if $a(t) = -1$, we have that
\begin{equation*}
a(t) - Q a(x) + \sum_{a(j) \geq a(i)}C_{ji}(a(j) - a(i))~\geq~-1 + \sum_{\substack{a(j) = 0 \\ a(i) = -1}}f_1(j,i)~=~-1 + 1~=~0,
\end{equation*}
since $(\forall i \neq j)~C_{ji} \geq f_1(j,i)$ and $f_1$ is an unit flow from $S$ to $t$.

Now assume $(\forall i)~a(i) \in \{0,1\}$, which defines a cut in G between nodes $\{i ~|~a(i) = 0\}$ (including $S$) and nodes $\{i~|~a(i) = 1\}$. If $a(t) = a(x) = 0$
\begin{equation*}
a(t) - Q a(x) + \sum_{a(j) \geq a(i)}C_{ji}(a(j) - a(i))~\geq~\sum_{\substack{a(j) = 1 \\ a(i) = 0}} f_1(j,i)~=~0,
\end{equation*}
since $(\forall i \neq j)~C_{ji} \geq f_1(j,i)$ and $S, x, t$ are all on the same side of the cut.
If $a(t) = 1, a(x) = 0$,
\begin{equation*}
a(t) - Q a(x) + \sum_{a(j) \geq a(i)}C_{ji}(a(j) - a(i))~\geq~1 + \sum_{\substack{a(j) = 1 \\ a(i) = 0}}f_2(j,i)~\geq~1 - 1~=~0,
\end{equation*}
since $(\forall i \neq j)~C_{ji} \geq f_2(j,i)$ and $f_2$ has a total flow value of $-1$ across any cut from $t$ to $S \cup \{x\}$. If $a(t) = a(x) = 1$,
\begin{equation*}
a(t) - Q a(x) + \sum_{a(j) \geq a(i)}C_{ji}(a(j) - a(i))~\geq~1 - Q + \sum_{\substack{a(j) = 1 \\ a(i) = 0}}f_2(j,i)~\geq~1 - Q - (1 - Q)~=~0,
\end{equation*}
since $(\forall i \neq j)~C_{ji} \geq f_2(j,i)$, $f_2$ has a total flow value of $-(1 - Q)$ across any cut between $S$ and $t$, and the flow from $x$ does not contribute to this cut, $x$ being on the same side as $t$. Finally if $a(t) = 0, a(x) = 1$,
\begin{equation*}
a(t) - Q a(x) + \sum_{a(j) \geq a(i)}C_{ji}(a(j) - a(i))~\geq~-Q + \sum_{\substack{a(j) = 1 \\ a(i) = 0}}f_2(j,i)~=~-Q + Q~=~0,
\end{equation*}
since $(\forall i \neq j)~C_{ji} \geq f_2(j,i)$ and $f_2$ has total flow $Q$ from $x$ to $t$, and $S$ does not contribute being on same side of the cut as $t$.
Hence~\eqref{e:gffsub9} is non-negative and the Lemma follows. 
\EPRF

\section{Gaussian MRF on Bounded Tree-width Graphs}
\label{sec:GPtreewidth}

In this section we will give an approximately optimal algorithm, to be precise a FPTAS, for Gaussian MRFs on bounded tree-width graphs. Consider a Gaussian MRF ${\bf X} = \big(X_1, X_2, \dots, X_n\big)$ on graph $G$ of tree-width at most $\kappa$. The bound on the tree-width implies that $\Lambda$ is sparse, having fewer than $\kappa n$ non-zero entries. Note that the precision matrix $\Lambda$ is defined only for non-degenerate Gaussians (i.e., with covariance matrix of full rank), and hence we will assume $\Lambda$ is of full rank. In this section, we present an approximation algorithm based on message passing for such a Gaussian MRF. We will show that for the special case of GFFs, the running time of this  message passing algorithm scales as $n^{O(\kappa^3)}$ in the number of variables.

We are going to use the notions of tree-decompositions and elimination orders associated with such decompositions, and refer to~\citet{Bod07} for a survey. Let $\bigg(\{V_1, V_2, \dots, V_m\},~T = \big(\{1,2,\dots,m\},F\big)\bigg)$ be any tree-decomposition (i.e., a junction tree) of $G$ of width $\kappa' \geq \kappa$. As we show later, using shallow tree-decompositions of width greater than the smallest possible (i.e., $\kappa$) may yield faster algorithms. Hence, each cluster $V_i \subseteq \{1, 2, \dots, n\}$ has size at most $\kappa' + 1$.

\BNTE
\label{note:decompassume}
We will assume w.l.o.g.
\begin{itemize}
\item that $m \geq n$  (using a bigger width $\kappa' > \kappa$ might lead to more than $n$ clusters),
\item that each non-leaf cluster in tree $T$ has degree $3$ ,
\item that $T$ has an empty cluster, $V_m = \emptyset$ (say), which is a leaf, and
\item that $1,2, \dots, n$ is an elimination order for the given tree-decomposition.
\end{itemize}
\ENTE

Such tree-decompositions exist---one can always transform a given tree-decomposition into a strictly binary tree (using, e.g., a trick like Figure~4.3 of~\citet{Bod88}) first and then add the empty cluster $V_m$ as the $3^{rd}$ neighbour of the ``root" (i.e. the cluster with exactly 2  neighbours). These assumptions about the tree-decomposition will help us give a clearer presentation of our algorithm.

Now each edge $\{i, j\} \in F$ splits the tree $T$ into 2 component subtrees $T_{ij}$ (containing cluster $i$) and $T_{ji}$ (containing cluster $j$) consisting respectively of the following nodes in $G$:
$$V_{ij}~\define \bigcup_{l \in T_{ij}}~V_l \quad \textrm{and} \quad V_{ji}~\define \bigcup_{l \in T_{ji}} V_l.$$ 

Now define the sets $\Gamma_{ij},~\Delta_{ij}$ as
$$\Delta_{ij}~\define~V_i \bigcap V_j~=V_{ij} \cap V_{ji} \quad \textrm{and} \quad \Gamma_{ij}~\define~V_i \setminus V_j.$$

The set $\Delta_{ij}$ separates nodes in $G$ into $V_{ij} \setminus \Delta_{ij}$ and $V_{ji} \setminus \Delta_{ij}$. This means the variance of any node in $V_{ji} \setminus \Delta_{ij}$ does not depend on which nodes in $V_{ij} \setminus \Delta_{ij}$ are observed, as long as we know the conditional joint distribution of variables in $\Delta_{ij}$ given these observations. For the case of Gaussian MRFs, the conditional distribution of variables in $\Delta_{ij}$ happens to be specified fully by the joint precision matrix. Intuitively, one can think of the observations in $V_{ij} \setminus \Delta_{ij}$ as inducing a ``prior" on the shared variables ${\bf X}_{\Delta_{ij}}$. This Markov property allows us to use a dynamic programming algorithm. 

For our message passing scheme to work, we need to factorize the joint density function of ${\bf X}$ into a product of densities, one for each set $V_i$, as follows. 

\BLEM
\label{l:gpdecomp}
There exist precision matrices $\prec_{V_1}, \prec_{V_2}, \dots, \prec_{V_m}$ which give the factorization
\begin{equation}
\label{e:gpdecomp2}
\exp\bigg(-\frac{1}{2}{\bf X}^t \Lambda {\bf X}\bigg)~=~\prod_{j=1}^{m}\exp\bigg(-\frac{1}{2}{\bf X}^t \prec_{V_j} {\bf X}\bigg),
\end{equation}
and which have the following properties. For each $j$, $\prec_{V_j}$ has support $V_j \times V_j$ (i.e., $\prec_{V_j} \in {\cal X}_+^{V_j \times V_j}$), has rank $|V_j|$, and
\begin{equation}
\label{e:gpdecomp1}
\frac{\lambda_{min}(\Lambda)}{m}~\leq~\lambda_{min}\big(\prec_{V_j}\big)~\leq~\lambda_{max}\big(\prec_{V_j}\big)~\leq~\lambda_{max}(\Lambda).
\end{equation}

Moreover, $\prec_{V_1}, \prec_{V_2}, \dots, \prec_{V_m}$ can be computed from the given tree-decomposition in time $O(m\kappa'^2)$.
\ELEM


\BPRF[of Lemma~\ref{l:gpdecomp}:]
Since $1,2, \dots, n$ are in elimination order, the Cholesky decomposition (see, e.g., Chapter~2.6 of~\citet{HJ85}) of $\prec - \lambda_{min}(\prec)I_n~=~U^tU$ has the following property. For each $i$ there exists some cluster $V_{j_i}$ in the given tree-decomposition such that the support of the $i^{th}$ row of $U$, $U[i,:]$,  is included in $V_{j_i}$, i.e., $\{l~|~U[i, l] \neq 0\} \subseteq V_{j_i}$. This means 
\begin{equation}
\label{e:gpdecomp3}
\begin{split}
{\bf X}^t \bigg( \prec - \lambda_{min}(\prec)I_n \bigg) {\bf X}~=~\sum_{i=1}^{n}{\bf X}^t U[i,:]^t U[i,:] {\bf X}~&=~\sum_{j=1}^m {\bf X}^t \bigg(~\sum_{i: j_i = j} U[i,:]^t U[i,:] \bigg) {\bf X} \\
~&=~\sum_{j=1}^m~{\bf X}^t \prec'_{V_j} {\bf X},
\end{split}
\end{equation}
where for each $j$, $\prec'_{V_j} \define \sum_{i: j_i = j} U[i,:]^t U[i,:]~\in~{\cal X}_+^{V_j \times V_j}$. Now notice that $I_n$ can be ``split" into diagonal matrices, $I_n~=~{\cal D}_{V_1} + {\cal D}_{V_2} + \dots + {\cal D}_{V_m}$, such that in each ${\cal D}_{V_j}$, each entry in the principal diagonal indexed by $V_j$ is at least $1/m$ and the rest of the entries are 0. Hence the matrices 
\begin{equation}
\label{e:gpdecomp4}
(\forall j) \quad \prec_{V_j}~\define~\prec'_{V_j}~+~\lambda_{min} {\cal D}_{V_j}
\end{equation}
satisfy the lower bound on eigenvalue in~\eqref{e:gpdecomp1}, i.e., $\frac{\lambda_{min}(\Lambda)}{m}~\leq~\lambda_{min}\big(\prec_{V_j}\big)$. The desired factorization~\eqref{e:gpdecomp2} as well as the upper bounds on eigenvalues in~\eqref{e:gpdecomp1}, i.e., $\lambda_{min}\big(\prec_{V_j}\big) \leq \lambda_{max}(\Lambda)$, now follow from combining~\eqref{e:gpdecomp3} and~\eqref{e:gpdecomp4}. \\
 
As for the time complexity, note that since $1,2, \dots, n$ are in elimination order, each step in the Cholesky decomposition algorithm takes only $O(\kappa'^2)$ time using sparse matrix representations and hence the total running time is $O(m\kappa'^2)$.
\EPRF

We will also make use of two transformations for precision matrices, $\mathrm{Obs}$ and $\mathrm{Marginal}$, which correspond respectively to observing some variables and computing the marginal over a subset of variables in each cluster. 

Given precision $\Lambda' \in {\cal X}_+^{V \times V}$ and a set of observed variables $O \subseteq V$, function $\mathrm{Obs}$ transforms $\Lambda'$ into a marginal precision matrix for variables ${\bf X}_{V \setminus O}$ by setting the rows and columns of $\Lambda'$ indexed by $O$ to $0$. In other words, if $\Lambda'' = \mathrm{Obs}\big(\Lambda',~O\big)$, then $\Lambda''$ has support $(V \setminus O) \times (V \setminus O)$ and 
\begin{equation}
\label{e:observe}
\Lambda''[V \setminus O,~V \setminus O]~\define~\Lambda'[V \setminus O,~V \setminus O].
\end{equation}

Transformation $\mathrm{Marginal}_{V, \Delta}\big(\Lambda'\big)$ computes the precision matrix of marginal distribution of variables in $\Delta$.  Transformation $\mathrm{Marginal}_{V, \Delta}\big(\Lambda'\big)$, where $\Delta \subseteq V$,  is defined only if $\Lambda[V \setminus \Delta, V \setminus \Delta]$ is of full rank.  $\mathrm{Marginal}_{V, \Delta}\big(\Lambda'\big)$ has support $\Delta \times \Delta$ and (see, e.g., Chapter~A.2 of~\citet{RW06}) 
\begin{equation}
\label{e:marginal}
\mathrm{Marginal}_{V, \Delta}\big(\Lambda'\big)[\Delta, \Delta]~=~\Lambda'[\Delta, \Delta]~-~\Lambda'[\Delta, V \setminus \Delta]\big(\Lambda'[V \setminus \Delta, V \setminus \Delta] \big)^{-1}\Lambda'[V \setminus \Delta, \Delta].
\end{equation}

In other words, $\mathrm{Marginal}_{V, \Delta}\big(\Lambda'\big)$ can be thought of as a sequence of 3 operations: first invert $\Lambda'[V, V]$, then take a principal submatrix (indexed by $V \setminus \Delta$) of the inverse, and finally invert the resulting submatrix.
Intuitively, $\mathrm{Marginal}_{V, \Delta}$ corresponds to ``integrating out" the other variables ${\bf X}_{V \setminus \Delta}$, and the following lemma makes this intuition precise.
\BLEM
\label{l:marginal}
For any $\Lambda' \in {\cal X}_+^{V \times V}$ and any $\Delta \subseteq V$, if $\mathrm{Marginal}_{V, \Delta}\big(\Lambda'\big)$ is defined, then
$$\int~\exp\bigg(-\frac{1}{2} {\bf X}^t \Lambda' {\bf X} \bigg)~\prod_{l \in V \setminus \Delta}\mathrm{d}X_l~~=~~C\big(\Lambda', \Delta\big)~\exp\bigg(-\frac{1}{2} {\bf X}^t~\mathrm{Marginal}_{V, \Delta}\big(\Lambda'\big)~{\bf X} \bigg),$$
where $C\big(\Lambda', \Delta\big)$ does not depend on ${\bf X}_{V \setminus \Delta}$. 
\ELEM
 
The complexity of computing $\mathrm{Marginal}$ and $\mathrm{Obs}$ are $O(|V|^3)$ and $O(|V|^2)$ respectively when the input matrix has  support $V \times V$. We note that application of $\mathrm{Marginal}$ or $\mathrm{Obs}$ does not make the smallest non-zero eigenvalue of the input matrix any smaller. This property will be useful later (Section~\ref{sec:GPtreewidthapprox}) when we discuss how to perform approximate message passing.

\BLEM
\label{l:eigenpreserve}
For any $\Lambda' \in {\cal X}_+^{V \times V}$ of rank $|V|$, any $O \subseteq V$ and any $\Delta \subseteq V$,
$$\lambda_{min}\big( \mathrm{Obs}\big(\Lambda',~O\big)\big) \geq \lambda_{min} \big( \Lambda'\big) \quad \textrm{and} \quad \lambda_{min} \big( \mathrm{Marginal}_{V, \Delta}\big(\Lambda'\big) \big) \geq \lambda_{min} \big( \Lambda'\big)$$
\ELEM

\BPRF[of Lemma~\ref{l:eigenpreserve}:]
That $\lambda_{min}\big( \mathrm{Obs}\big(\Lambda',~O\big)\big) \geq \lambda_{min} \big( \Lambda'\big)$ follows from the fact that  taking a principal submatrix of a positive definite matrix (more generally, of any symmetric matrix), in this case $\Lambda'[V, V]$, does not decrease the smallest eigenvalue. 
Similarly, $\mathrm{Marginal}_{V, \Delta}\big(\Lambda'\big)$ consists first inverting $\Lambda'[V, V]$, then taking a principal submatrix indexed by $V \setminus \Delta$, and finally inverting the resulting submatrix. The largest eigenvalue of the inverse, $(\Lambda'[V, V])^{-1}$, is at most $\lambda_{min}\big((\Lambda'[V, V])\big)^{-1}$, which can only decrease after taking the submatrix in the second step. Hence after the final inversion in the third step, the smallest eigenvalue is at least $\lambda_{min}\big(\Lambda'[V, V]\big)$, which equals $\lambda_{min}\big(\Lambda'\big)$ since $\Lambda'$ is of rank $|V|$ and has support $V \times V$. 
\EPRF

For any set $\Delta \subseteq V_{i}$ and a set of observations $S \subseteq V_{ij}$, we will say that variables ${\bf X}_{\Delta}$ \emph{have precision} $P$ \emph{given} $S$ if $P$ is the precision matrix of the ``prior" induced by observations $S$ on variables ${\bf X}_{\Delta}$, considering only those factors in Lemma~\ref{l:gpdecomp} which belong to the subtree $T_{ij}$  in the given tree-decomposition. 

\BDEF
Consider any edge $\{i, j\} \in F$ in the tree-decomposition. Then for any set $\Delta \subseteq V_{i}$ and any set of observations $O \subseteq V_{ij}$, we say ${\bf X}_{\Delta}$ \emph{has precision} $P$ \emph{given} $O$ \emph{in} $T_{ij}$ if 
$$P = \mathrm{Marginal}_{V_{ij} \setminus O, \Delta \setminus O}\bigg(\mathrm{Obs}\bigg(\sum_{l \in T_{ij}} \Lambda_{V_l},~O\bigg)\bigg).$$
\EDEF

Note that considering only factors which belong to subtree $T_{ij}$ defines a different set of random variables with a different density than the original set, i.e., ${\bf X} = \big(X_1, X_2, \dots, X_n\big)$; however, we will slightly abuse the notation and still use ${\bf X}$, and the actual distribution of these random variables should be clear from the context. We also need to introduce the following notation which describes the total error achieved in a subtree of the given tree-decomposition given observations in that subtree. 

\BDEF
For each edge $\{i, j\} \in F$, for any precision matrix $Q \in {\cal X}_+^{\Delta_{ij} \times \Delta_{ij}}$ consider the (origin-centred) Gaussian density $\rho_{ij, Q}$ on variables ${\bf X}_{V_{ij}}$ in subtree $T_{ij}$ defined by the precision matrix $Q + \sum_{l \in T_{ij}} \Lambda_{V_l}$. Then for any set $O \subseteq V_{ij}$ of observed variables,
\begin{equation}
\label{e:approxerr}
\begin{split}
R_{ij}\big(Q, O\big)~&\define~\sum_{t \in V_{ij} \setminus \Delta_{ij}}~\var_{\rho_{ij, Q}}\big[X_t~|~{\bf X}_O\big].
\end{split}
\end{equation}
In other words, $R_{ij}\big(Q, O\big)$ is the total error of variables in $V_{ij} \setminus \Delta_{ij}$ when ${\bf X}_{\Delta_{ij}}$ has $Q$ as a ``prior" due to observations which lie outside $V_{ij}$.
\EDEF

We are now going to describe an idealized message passing algorithm which finds the exact optimum of the budget version, but uses messages that are functions on continuous domains. Later we will show how to round the messages (making their size polynomial) at the cost of producing an approximate solution.

Intuitively, the message is a function that gives the optimal total error in one part of the graph ($V_{ij}$) for every possible way of splitting the budget between the parts ($V_{ij}\setminus\Delta_{ij},V_{ji}\setminus\Delta_{ij},\Delta_{ij}$), for every possible choice of observations in $\Delta_{ij}$ (respecting the budget allocation), and for every possible pair of distributions of the shared variables ($\Delta_{ij}$) where the first distribution comes from the Gaussian MRF on $V_{ij}$ (that is, using only the factors in~\eqref{e:gpdecomp2} that are in $V_{ij}$) and the second distribution comes  from the Gaussian MRF on $V_{ji}$ (again, using only the factors in~\eqref{e:gpdecomp2} that are in $V_{ji} \setminus \Delta_{ij}$). Note that we allow the allotted number of observations for first (and also for the second) distribution but their location is not communicated in the message (this is justified by the Markov property discussed earlier).

The message passing algorithm proceeds in a sequence of rounds. In  each round, cluster $i$ in $T$ optionally sends a message $\alpha_{i \to j}$ to its  neighbouring cluster $j$ along edge $\{i,j\} \in F$. In the first round, each leaf in $T$ sends a message to its (only) neighbour. Once $i$ has received a message from each of its neighbours excluding $j$, $i$ sends a message to $j$ exactly once in the following round. The message $\alpha_{i \to j}$ is a function
\begin{equation*}
\begin{split}
\alpha_{i \to j} :~{\cal X}_+^{\Delta_{ij} \times \Delta_{ij}} \times {\cal X}_+^{\Delta_{ij} \times \Delta_{ij}} \times 2^{\Delta_{ij}} \times \{0, 1, 2, \dots, b\}~\to~\R,
\end{split}
\end{equation*}
interpreted as follows.  Given $Q, P \in {\cal X}^{\Delta_{ij} \times \Delta_{ij}}$, $S \subseteq \Delta_{ij}$ and $N \leq b$,  let ${\cal S}^*(P, S, N)$ be the collection of all sets $S' \subseteq V_{ij}$, each of which have the property
\begin{itemize}
\item that $S' \cap \Delta_{ij} = S$ (the variables in $S$ are the only ones to be observed among $\Delta_{ij}$),
\item that $|S'| \leq N$ (at most $N$ observations are allowed in $V_{ij}$), and
\item that ${\bf X}_{\Delta_{ij}}$ has precision $P$ given (only) observations $S'$ in subtree $T_{ij}$.
\end{itemize}
If ${\cal S}^*(P, S, N)$ is non-empty, then $\alpha_{i \to j}\big(P, Q, S, N\big)$ is defined as
\begin{equation}
\alpha_{i \to j}\big(P, Q, S, N\big) = \min_{\substack{S' \in {\cal S}^*(P, S, N),~S'' \subseteq V_{ji} \setminus \Delta_{ij} \\ {\bf X}_{\Delta_{ij}}~\textrm{has precision} \\ Q~\textrm{given}~S''~\textrm{in}~T_{ji}}} \sum_{t \in V_{ij} \setminus \Delta_{ij}} \var\big[X_t~\big|~{\bf X}_{S' \cup S''}\big], 
\end{equation}

Otherwise, $\alpha_{i \to j}\big(P, Q, S, N\big) = +\infty$. We point out that by~\eqref{e:approxerr}, one can also express 
$\alpha_{i \to j}$ as
$$\alpha_{i \to j}\big(P, Q, S, N\big)  = \min_{S' \in {\cal S}^*(P, S, N)}~R_{ij}\big(Q, S'\big).$$

It will be useful later to have a notion of the height of a message, defined as one plus the height of the maximum of the heights of messages from which it was composed, with the messages sent from leaves having a height of $1$. 

Given the definition of $\alpha_{i \to j}$ above, its value when cluster $V_i$ happens to be a leaf (i.e., when $\alpha_{i \to j}$ has height $1$) for arguments $P_{ij}, Q_{ji} \in {\cal X}^{\Delta_{ij} \times \Delta_{ij}}$, $S_{ij} \subseteq \Delta_{ij}$ and $N_i$ can be expressed as
\begin{equation}
\label{e:gpleaf}
\begin{split}
\alpha_{i \to j}\big(P_{ij}, Q_{ji}, S_{ij}, N_i \big) &= 
\underset{\substack{L_{ij} \subseteq \Gamma_{ij} \\ |L_{ij}| + |S_{ij}| \leq N_i}}{\min}~\TR\bigg( \big(\Lambda'_{V_{i}}\big[V'_i, V'_i\big]\big)^{-1} \bigg), \quad \textrm{with} \\
V'_i =  \Gamma_{ij} \setminus L_{ij}~&~\textrm{and}~\Lambda'_{V_i} = \mathrm{Obs}\big(\Lambda_{V_i} + Q_{ji},~S_{ij} \cup L_{ij} \big),
\end{split}
\end{equation}
where the set $L_{ij}$ in~\eqref{e:gpleaf} must satisfy the additional constraint that  ${\bf X}_{\Delta_{ij}}$ has precision $P_{ij}$ given observations $S_{ij} \cup L_{ij}$ in $T_{ij}$.

Next we describe how the messages are composed for internal nodes. Suppose cluster $i$ has 3 neighbours $j, k$ and $l$, and consider the earliest round by which $i$ has received messages $\alpha_{k \to i}$ and $\alpha_{l \to i}$ from clusters $k$ and $l$ respectively. Then the function $\alpha_{i \to j}$ for arguments $P_{ij}, Q_{ji}$, $S_{ij}, N_i$ is composed in the following round recursively using~\eqref{e:compose'}-\eqref{e:compose3} as follows. 


\begin{equation}
\label{e:compose'}
\begin{split}
\alpha_{i \to j} \big(P_{ij}, Q_{ji}, S_{ij}, N_i \big) = \underset{\substack{P_{ki}, Q_{ik}, S_{ik}, N_k \\ P_{li}, Q_{il}, S_{il}, N_l, L_{ij}}}{\min} 
 & \alpha_{k \to i} \big(P_{ki}, Q_{ik}, S_{ik}, N_k \big) + \alpha_{l \to i}\big(P_{li}, Q_{il}, S_{il}, N_l \big)  \\
& \quad + \TR\bigg( \big({\Lambda'}_{V_{i}}\big[V'_i, V'_i\big] \big)^{-1} \bigg),
 \end{split}
\end{equation} 
with 
$$\qquad V'_i = \Gamma_{ij} \setminus L_{ij},~\textrm{and}~\Lambda'_{V_{i}} = \mathrm{Obs}\bigg(\Lambda_{V_i} + P_{li} + P_{ki} + Q_{ji},~S_{ij} \cup L_{ij}\bigg),$$
and where the minimum (infimum) in~\eqref{e:compose'} is taken over all $L_{ij} \subseteq \Gamma_{ij}$ and over all $P_{ki}, Q_{ik}, S_{ik}, N_k$ and $P_{li}, Q_{il}, S_{il}, N_l$ satisfying~\eqref{e:compose2},~\eqref{e:compose1} and~\eqref{e:compose3} below: 
\begin{equation}
\label{e:compose2}
\begin{split}
S_{ik} = (S_{ij} \cup L_{ij}) \cap \Delta_{ik},~S_{il} = (S_{ij} \cup L_{ij}) \cap \Delta_{il},~|S_{ij}| + |L_{ij}| + N_k + N_l - |S_{ik}| - |S_{il}| \leq N_i,
\end{split}
\end{equation}

\begin{equation}
\label{e:compose1}
P_{ij}~=~\mathrm{Marginal}_{V_i \setminus (S_{ij} \cup L_{ij}),~\Delta_{ij} \setminus S_{ij}}\bigg(\mathrm{Obs}\bigg(\Lambda_{V_i} + P_{ki} + P_{li},~S_{ij} \cup L_{ij} \bigg) \bigg),
\end{equation}

and further,
\begin{equation}
\label{e:compose3}
\begin{split}
Q_{ik} &= \mathrm{Marginal}_{V_i \setminus (S_{ij} \cup L_{ij}),~\Delta_{ik} \setminus S_{ik}}\bigg(\mathrm{Obs}\bigg( \Lambda_{V_i} + P_{li}  +Q_{ji},~S_{ij} \cup L_{ij} \bigg) \bigg), \\
Q_{il} &= \mathrm{Marginal}_{V_i \setminus (S_{ij} \cup L_{ij}),~\Delta_{il} \setminus S_{il}}\bigg(\mathrm{Obs}\bigg( \Lambda_{V_i} + P_{ki} + Q_{ji},~S_{ij} \cup L_{ij} \bigg) \bigg).
\end{split}
\end{equation} 

We will use $P_{ki}^*,Q_{ik}^*, S_{ik}^*, N_k^*$, $P_{li}^*, Q_{il}^*, S_{il}^*$ and $N_l^*$ to denote the values of 
$P_{ki}, Q_{ik}, S_{ik}, N_k$, $P_{li}, Q_{il}, S_{il}, N_l$ respectively for which the minimum  in~\eqref{e:compose'} is achieved.

 Once we have the ideal message passing algorithm, it is easy to describe how the approximate messages are composed. Assume that there is a transformation $\mathrm{Round}_{\eps}$ which, for any $\eps > 0$, $\eps$-approximates any precision matrix with support $V \times V$ by mapping it to an element of an $\eps$-net, ${\cal I}_{\eps}^{V \times V}$, for such matrices. We defer the precise definitions of our notion of $\eps$-approximation, the transformation $\mathrm{Round}_{\eps}$ and the $\eps$-nets ${\cal I}_{\eps}^{V \times V}$ until later. The approximate messages will be only defined for precision matrices in ${\cal I}_{\eps}^{V \times V}$ where $V \subseteq \Delta_{ij}$, and we will only require~\eqref{e:compose1} and~\eqref{e:compose3} to hold approximately. To be precise given arguments $\hat{P}_{ij}, \hat{Q}_{ji}, \hat{S}_{ij}, \hat{N}_i$, we have

\begin{equation}
\label{e:acompose'}
\begin{split}
\hat{\alpha}_{i \to j}\big(\hat{P}_{ij}, \hat{Q}_{ji}, \hat{S}_{ij}, \hat{N}_i \big) = \underset{\substack{\hat{P}_{ki}, \hat{Q}_{ik}, \hat{S}_{ik}, \hat{N}_k \\ \hat{P}_{li}, \hat{Q}_{il}, \hat{S}_{il}, \hat{N}_l, \hat{L}_{ij}}}{\min} & \hat{\alpha}_{k \to i} \big(\hat{P}_{ki}, \hat{Q}_{ik}, \hat{S}_{ik}, \hat{N}_k \big) + \hat{\alpha}_{l \to i} \big(\hat{P}_{li}, \hat{Q}_{il}, \hat{S}_{il}, \hat{N}_l \big) \\ 
 & \quad + \TR \bigg( \big({\hat{\Lambda}'}_{V_{i}} \big[\hat{V}'_i, \hat{V}'_i\big] \big)^{-1} \bigg),
\end{split}
\end{equation}
with
$$ \hat{V}'_i = \Gamma_{ij} \setminus \hat{L}_{ij},~\textrm{and}~{\hat{\Lambda}}'_{V_i} = \mathrm{Obs}\bigg(\Lambda_{V_i} + \hat{P}_{li} + \hat{P}_{ki} + \hat{Q}_{ji},~\hat{S}_{ij} \cup \hat{L}_{ij} \bigg),$$
and where the minimum (infimum) in~\eqref{e:acompose'} is taken over all $\hat{L}_{ij} \subseteq \Gamma_{ij}$ and all $\hat{P}_{ki}, \hat{Q}_{ik}, \hat{S}_{ik}, \hat{N}_k$ and $\hat{P}_{li}, \hat{Q}_{il}, \hat{S}_{il}, \hat{N}_l$ such that $\hat{P}_{ki}, \hat{Q}_{ik} \in {\cal I}_{\eps}^{\Delta_{ik} \setminus \hat{S}_{ik} \times \Delta_{ik} \setminus \hat{S}_{ik}},~\hat{P}_{li}, \hat{Q}_{il} \in {\cal I}_{\eps}^{\Delta_{il} \setminus \hat{S}_{il} \times \Delta_{il} \setminus \hat{S}_{il}}$, and 
which satisfy~\eqref{e:acompose2},~\eqref{e:acompose1} and~\eqref{e:acompose3} below:

\begin{equation}
\label{e:acompose2}
\begin{split}
\hat{S}_{ik} = \big(\hat{S}_{ij} \cup \hat{L}_{ij} \big) \cap \Delta_{ik},~\hat{S}_{il} = \big(\hat{S}_{ij} \cup \hat{L}_{ij} \big) \cap \Delta_{il},~|\hat{S}_{ij}| + |\hat{L}_{ij}| + \hat{N}_k + & \hat{N}_l - |\hat{S}_{ik}| - |\hat{S}_{il}| \leq \hat{N}_i,
\end{split}
\end{equation}


\begin{equation}
\label{e:acompose1}
\begin{split}
\hat{P}_{ij} = \mathrm{Round}_\eps\bigg(\mathrm{Marginal}_{V_i \setminus (\hat{S}_{ij} \cup \hat{L}_{ij}),~\Delta_{ij} \setminus \hat{S}_{ij}}\bigg(\mathrm{Obs}\big(\Lambda_{V_i} + \hat{P}_{ki} + \hat{P}_{li},~\hat{S}_{ij} \cup \hat{L}_{ij} \bigg) \bigg)\bigg),
\end{split}
\end{equation}
and
\begin{equation}
\label{e:acompose3}
\begin{split}
\hat{Q}_{ik} & = \mathrm{Round}_{\eps}\bigg( \mathrm{Marginal}_{V_i \setminus (\hat{S}_{ij} \cup \hat{L}_{ij}),~\Delta_{ik} \setminus \hat{S}_{ik}} \bigg(\mathrm{Obs}\bigg( \Lambda_{V_i} + \hat{P}_{li}  +\hat{Q}_{ji},~\hat{S}_{ij} \cup \hat{L}_{ij} \bigg) \bigg) \bigg), \\
\hat{Q}_{il} & = \mathrm{Round}_{\eps}\bigg( \mathrm{Marginal}_{V_i \setminus (\hat{S}_{ij} \cup \hat{L}_{ij}),~\Delta_{il} \setminus \hat{S}_{il}} \bigg(\mathrm{Obs}\bigg( \Lambda_{V_i} + \hat{P}_{ki} + \hat{Q}_{ji},~\hat{S}_{ij} \cup \hat{L}_{ij} \bigg) \bigg) \bigg).
\end{split}
\end{equation}

We will use $\hat{P}_{ki}^*,\hat{Q}_{ik}^*, \hat{S}_{ik}^*, \hat{N}_k^*$, $\hat{P}_{li}^*, \hat{Q}_{il}^*, \hat{S}_{il}^*, \hat{N}_l^*$ and  $\hat{L}_{ij}^*$ to denote  arguments for which minimum is achieved~\eqref{e:acompose'}.

Finally, an approximate message $\hat{\alpha}_{i \to j}$ sent from a leaf cluster $V_i$ is given by the same equation~\eqref{e:gpleaf} as the ideal message, except that the joint precision matrix of ${\bf X}_{\Delta_{ij}}$ is rounded to ${\cal I}_{\eps}^{(\Delta_{ij} \setminus \hat{S}_{ij}) \times (\Delta_{ij} \setminus \hat{S}_{ij})}$, i.e.,

\begin{equation}
\label{e:agpleaf}
\begin{split}
\alpha_{i \to j}\big(\hat{P}_{ij}, \hat{Q}_{ji}, \hat{S}_{ij}, \hat{N}_i \big) &= 
\underset{\substack{ \hat{L}_{ij} \subseteq \Gamma_{ij} \\ |\hat{S}_{ij}| + |\hat{L}_{ij}| \leq \hat{N}_i}}{\min}~\TR\bigg( \big(\hat{\Lambda}'_{V_i}[\hat{V}'_i, \hat{V}'_i]\big)^{-1} \bigg), \quad \textrm{with} \\
\hat{V}'_i = \Gamma_{ij} \setminus \hat{L}_{ij}, &~\textrm{and}~\hat{\Lambda}'_{V_i} = \mathrm{Obs}\big(\Lambda_{V_{i}} + \hat{Q}_{ji},~\hat{S}_{ij} \cup \hat{L}_{ij}\big),
\end{split}
\end{equation}
where the set $\hat{L}_{ij}$ in~\eqref{e:agpleaf} must satisfy the additional constraint that the precision $P$ of ${\bf X}_{\Delta_{ij}}$ given observations $\hat{S}_{ij} \cup \hat{L}_{ij}$ in tree $T_{ij}$ is such that  $\hat{P}_{ij}~=~\mathrm{Round}_{\eps}(P)$.

Now that we have described how the approximate messages are composed, we describe how to construct the $\eps$-nets ${\cal I}_{\eps}^{V \times V}$ and the transformation $\mathrm{Round}_{\eps}$ in the following subsections. 

\subsection{Approximate Message Passing for GFFs}
\label{sec:gffround}

For the special case of a GFF~\eqref{e:gff} (i.e., when  the precision matrix $\Lambda$ is a graph Laplacian~\eqref{e:gffLaplacian}) on bounded tree-width graphs, we can use a simple notion of approximation which rounds off each element in the precision matrices being passed. Before we analyze this approximation scheme, observe that given our tree-decomposition, one can split $\Lambda$ as $\Lambda = \sum_i~\Lambda_{V_i}$ simply by defining $\Lambda_{V_i}$ for each $i$ to be the Laplacian of a subgraph induced by vertices in cluster $V_i$, with the understanding that each edge shared by 2 or more clusters is assigned to the subgraph induced by exactly one of these clusters. Unlike Lemma~\ref{l:gpdecomp}, this split does not use Cholesky decomposition.

\BOBS
\label{o:gffdecomp}
For a GFF, there exist precision matrices $\prec_{V_1}, \prec_{V_2}, \dots, \prec_{V_m}~\in~{\cal G}^{n \times n}$ such that each $\prec_{V_j}$ has support $V_j \times V_j$, and which give the factorization
\begin{equation*}
\exp\bigg(-\frac{1}{2}{\bf X}^t \Lambda {\bf X}\bigg)~=~\prod_{j=1}^{m}\exp\bigg(-\frac{1}{2}{\bf X}^t \prec_{V_j} {\bf X}\bigg).
\end{equation*}
The matrices $\prec_{V_1}, \prec_{V_2}, \dots, \prec_{V_m}$ can be computed in time $O(m\kappa)$ using sparse representations.
\EOBS

In fact for GFFs all the precision matrices obtained during message passing are going to be symmetric diagonally dominant with non-positive off-diagonal entries, i.e., from ${\cal G}^{n \times n}$. 

\BDEF
\label{e:gffrel} 
Given 2 matrices $Q, Q' \in {\cal G}^{n \times n}$ and any $\eps > 0$, we will say that $Q \approxeq_{\eps} Q'$ if 
\begin{equation*}
\begin{split}
(\forall i \neq j) \quad &\e^{-\eps} \big|Q[i, j]\big|~\leq~\big|Q'[i, j]\big|~\leq~\e^{\eps} \big|Q[i, j]\big|,~\textrm{and} \\
(\forall i) \quad \quad &\e^{-\eps} \sum_{j} Q[i, j]~\leq~\sum_{j} Q'[i, j]~\leq~\e^{\eps} \sum_{j} Q[i, j].
\end{split}
\end{equation*}
\EDEF

Note that the definition requires that the respective row sums, which are non-negative since matrices in ${\cal G}^{n \times n}$ are diagonally dominant, rather than the diagonal elements, be approximately equal. It is easy to verify that $\approxeq$ satisfies the following ``triangle inequality".

\BOBS
\label{o:trianglegff}
If $Q, Q_1, Q_2 \in {\cal G}^{n \times n}$ and $\eps_1, \eps_2 \geq 0$ are such that $Q_1 \approxeq_{\eps_1} Q$ and $Q_2 \approxeq_{\eps_2} Q$, then $Q_1 \approxeq_{\eps_1 + \eps_2} Q_2$. 
\EOBS

We approximate the precision matrices obtained while message passing by rounding the off-diagonal elements and the row sums, for which we need to calculate the range of values these elements can assume. Each non-zero element (or row sum) of each precision matrix obtained while running the ideal message passing algorithm~\eqref{e:compose'}-\eqref{e:compose3} for a GFF lies in the range $[c_l, c_h]$, with $c_l, c_h$ as defined in \eqref{e:gffepsnet} below. Intuitively, $c_h$ is the 
largest possible value of the effective conductance (i.e., inverse of the effective resistance) between any 2 nodes in the electrical network associated with the GFF. Similarly $c_l$ is, roughly speaking, the smallest possible effective conductance between 2 nodes. Our $\eps$-nets for GFFs are going to be the following subsets of ${\cal G}^{n \times n}$.
\begin{equation}
\label{e:gffepsnet}
\begin{split}
{\cal I}_{\eps}^{V \times V}~&\define~\bigg\{ P \in {\cal G}^{n \times n}~\big|~P~\textrm{has support}~V \times V~\textrm{and}~(\forall i \neq j)~|P[i, j]| \in {\cal L}_{\eps}~\textrm{and} \\
~& \qquad \qquad \qquad \qquad (\forall i)~\bigg(\sum_j P[i, j]\bigg) \in {\cal L}_{\eps} \bigg\},\\
\textrm{where} & ~\\
{\cal L}_{\eps}~&\define~\bigg\{0\bigg\} \bigcup \bigg\{c_l,~\e^{\eps}c_l,~\e^{2\eps} c_l, \dots,~\e^{\lfloor \frac{\ln (c_h / c_l)}{\eps} \rfloor \eps} c_l\bigg\}, \qquad \textrm{and where}\\
c_l~&=~\frac{\min_{\Lambda[i, j] > 0} |\Lambda[i, j]|^2}{n \max_{i \neq j} |\Lambda[i, j]|} = \frac{\min_{i \neq j} r_{ij}}{n\max_{ \{i, j\} \in E} r_{ij}^2}, \\
c_h~&=~\frac{n\max_{i \neq j} |\Lambda[i, j]|}{2} = \frac{n}{2\min_{i \neq j} r_{ij}}.
\end{split}
\end{equation}

\BOBS
\label{o:gffepsnetsize}
For any $\eps > 0$ and set $V$, the size of ${\cal I}_{\eps}^{V \times V}$ is bounded as
$$\big| {\cal I}_{\eps}^{V \times V} \big|~\leq~\bigg(2 + \frac{1}{\eps}\ln\bigg( \frac{n^2}{2} \frac{\max_{ \{i, j\} \in E} r_{ij}^2}{\min_{i \neq j} r_{ij}^2} \bigg) \bigg)^{|V|^2}.$$
\EOBS

We are now ready to define our transformation $\mathrm{Round}_{\eps}$. Once again, note that we round-off the row sums instead of the diagonal elements. We point out that our definition of $\mathrm{Round}_{\eps}$ is such that each off-diagonal element and each row sum of each precision matrices obtained during approximate message passing always stay within the range $[c_l, c_h]$ (with $c_l, c_h$ as defined in \eqref{e:gffepsnet}). For any $\eps > 0$ and for any $P \in {\cal G}^{n \times n}$, $P' = \mathrm{Round}_{\eps}(P)$ is given by
\begin{equation}
\label{e:gffround}
\begin{split}
(\forall i \neq j) \quad P'[i, j]~&\define~-\arg\underset{r \in {\cal L}_{\eps}}{\min}~\big|r - |P[i, j]|\big|, \quad \textrm{and} \\
(\forall i) \quad P'[i, i]~&\define~\arg \underset{r \in {\cal L}_{\eps}}{\min} \big|r - \sum_{j} P[i, j]\big| +  \sum_{j \neq i} \big| P'[i, j] \big|.
\end{split}
\end{equation}

$\mathrm{Round}_{\eps}$ has the following property.

\BLEM
\label{l:gffround}
Consider any $\eps > 0$ and any precision matrix $P \in {\cal G}^{n \times n}$ with support (say) $V \times V$ obtained while running the approximate message passing algorithm (given by~\eqref{e:acompose'}-\eqref{e:acompose3} and~\eqref{e:agpleaf}) for the given GFF. Then $\mathrm{Round}_{\eps}(P)  \in {\cal I}_{\eps}^{V \times V}$ and $P \approxeq_{\eps} \mathrm{Round}_{\eps}(P)$.
\ELEM

\BPRF
The proof of $P \approxeq_{\eps} \mathrm{Round}_{\eps}(P)$ is straightforward given the definition~\eqref{e:gffrel} of $\approxeq_{\eps}$ and the definition of the $\eps$-net ${\cal I}_{\eps}^{V \times V}$.
\EPRF

We next state and prove some useful properties of $\approxeq_{\eps}$ which will help us analyze how error introduced by rounding accumulates during message passing. Recall that our message passing algorithm (given by~\eqref{e:acompose'}-\eqref{e:acompose3}) makes use of the following operations and transformations on matrices: addition, $\mathrm{Obs}$ (which is equivalent to taking a principal submatrix), $\mathrm{Marginal}$, $\mathrm{Round}_{\eps}$ and finally, the trace of the inverse. Lemma~\ref{l:gffround} tells us how much error $\mathrm{Round}_{\eps}$ introduces. We analyze how each of the other transformations affect the rounding error.

\BLEM
\label{l:gffops}
Consider any $\eps \geq 0$. Then for any set $V$ and any $Q, Q' \in {\cal G}^{n \times n}$ with support $V \times V$ such that $Q'~\approxeq_{\eps}~Q$, we have 
\begin{equation}
\label{e:obsapxgff}
(\forall O \subseteq V) \quad \mathrm{Obs}\big(Q' ,~O\big)~\approxeq_{\eps}~\mathrm{Obs}\big(Q,~O\big), \quad \textrm{and}
\end{equation}
\begin{equation}
\label{e:margapxgff}
(\forall \Delta \subseteq V) \quad \mathrm{Marginal}_{V, \Delta}\big(Q'\big)~\approxeq_{\eps'}~\mathrm{Marginal}_{V, \Delta}\big(Q\big),~\textrm{where}~\eps' = 3^{|V \setminus \Delta|} \eps.
\end{equation}
Further, for any $Q_1, Q_2$ and $Q'_1, Q'_2$ in ${\cal G}^{n \times n}$ such that $Q'_1 \approxeq_{\eps} Q_1$ and $Q'_2 \approxeq_{\eps} Q_2$, we have
\begin{equation}
\label{e:addapxgff}
Q'_1 + Q'_2~\approxeq_{\eps}~Q_1 + Q_2.
\end{equation}
\ELEM

\BLEM
\label{l:traceapxgff}
If $Q, Q' \in {\cal G}^{n \times n}$ are such that $Q, Q'$ have support $V \times V$, $Q'~\approxeq_{\eps}~Q$, and $Q[V, V]$ has full rank (i.e., $|V|$), then 
$$\e^{-\eps}~\TR\big(Q[V, V]^{-1}\big)~\leq~\TR\big(Q'[V, V]^{-1}\big)~\leq~\e^{\eps}~\TR\big(Q[V, V]^{-1}\big).$$
\ELEM

Proofs of~\eqref{e:obsapxgff} and~\eqref{e:addapxgff} are straightforward; we prove Lemma~\ref{l:traceapxgff} and~\eqref{e:margapxgff} below.

\BPRF[of~\eqref{e:margapxgff}:]
We are going to prove the case where $|V \setminus \Delta| = 1$, i.e., when exactly 1 variable is being ``integrated out"; the general case follows readily by induction on $|V \setminus \Delta|$. Assume that $X_k, k \in V$ is being eliminated so that $\Delta = V \setminus \{k\}$, and let  $P = \mathrm{Marginal}_{V, \Delta}\big(Q\big),~P' = \mathrm{Marginal}_{V, \Delta}\big(Q'\big)$. By integrating the joint density of ${\bf X}_V$, or to be precise the function $\exp \big(-\frac{1}{2} {\bf X}^t Q {\bf X} \big)$, w.r.t. $X_k$ we get the following identity.
\begin{equation}
\label{e:margapxgff1}
\begin{split}
(\forall i \neq j) & \quad |P[i, j]|~=~|Q[i, j]|~+~\frac{|Q[i, k]| |Q[j, k]|}{Q[k, k]},~\textrm{and} \\
(\forall i) \quad & \sum_j P[i, j]~=~\sum_j Q[i, j]~+~\frac{|Q[i, k]| \sum_j Q[k, j]}{Q[k, k]}.
\end{split}
\end{equation}
Similarly, we have by integrating $\exp \big(-\frac{1}{2} {\bf X}^t Q' {\bf X} \big)$ w.r.t. $X_k$ that 
\begin{equation}
\label{e:margapxgff2}
\begin{split}
(\forall i \neq j) & \quad |P'[i, j]|~=~|Q'[i, j]|~+~\frac{|Q'[i, k]| |Q'[j, k]|}{Q'[k, k]},~\textrm{and} \\
(\forall i) \quad & \sum_j P'[i, j]~=~\sum_j Q'[i, j] ~+~\frac{|Q'[i, k]| \sum_j Q'[k, j]}{Q'[k, k]}.
\end{split}
\end{equation}
Taking ratios of the l.h.s. and r.h.s. of~\eqref{e:margapxgff1} and~\eqref{e:margapxgff2} and by the hypothesis that $Q'~\approxeq_{\eps}~Q$, we obtain
\begin{equation*}
\begin{split}
(\forall i \neq j) \quad &\e^{-3\eps} \big|P[i, j]\big|~\leq~\big|P'[i, j]\big|~\leq~\e^{3\eps} \big|P[i, j]\big|,~\textrm{and} \\
(\forall i) \quad \quad &\e^{-3\eps} \sum_{j} P[i, j]~\leq~\sum_{j} P'[i, j]~\leq~\e^{3\eps} \sum_{j} P[i, j],
\end{split}
\end{equation*}
which means $P \approxeq_{3\eps} P'$.
\EPRF

\BPRF[of Lemma~\ref{l:traceapxgff}:]
We are going to use the connection between electrical networks and GFFs, specifically Thomson's Principle, i.e., Lemma~\ref{l:thomson}. Note that any $Q, Q' \in {\cal G}^{n \times n}$ can be thought of as $n \times n$ principal submatrices of $(n+1) \times (n+1)$ graph Laplacians, respectively $P$ and $P'$, whose $(n+1)^{th}$ rows (and columns) are defined as 
\begin{equation*} 
\begin{split}
(\forall i \leq n) \quad &P[n+1, i] = -\sum_j Q[j, i],~~P'[n+1, i] = -\sum_j Q'[j, i], \\
\textrm{and} \quad & P[n+1, n+1] =  \sum_{j \leq n} | P[n+1, j] |,~~P'[n+1, n+1] =  \sum_{j \leq n} | P'[n+1, j] |.
\end{split}
\end{equation*}
Intuitively, one can think of $P, P'$ as GFFs with an additional variable $X_{n+1}$, in which case $\TR\big(Q[V, V]^{-1}\big), \TR\big(Q'[V, V]^{-1}\big)$ are simply the sum of conditional variances of each $X_l,~l \in V$, given $X_{n+1}$. However, by Thomson's principle we can express the conditional variance of $X_l$ given $X_{n+1}$ as the minimum energy of a unit flow from $n+1$ to $l$. To be precise, we have by Lemma~\ref{l:thomson} that 
\begin{equation}
\label{e:traceapxgff1}
\TR\big(Q[V, V]^{-1}\big)~=~\sum_{l \in V} \min_{f_l}~\sum_{i \neq j} \frac{f_l(i, j)^2}{|P[i, j]|},
\end{equation}
where each $f_l$ is constrained to be a unit flow from node $n+1$ to node $l$. Similarly, 
\begin{equation}
\label{e:traceapxgff2}
\TR\big(Q'[V, V]^{-1}\big)~=~\sum_{l \in V} \min_{f'_l}~\sum_{i \neq j} \frac{f'_l(i, j)^2}{|P'[i, j]|},
\end{equation}
where each $f'_l$ is is constrained to be a unit flow from node $n+1$ to node $l$. Now note that since $Q \approxeq_{\eps} Q'$, we have by definition of $P, P'$ that
\begin{equation}
\label{e:traceapxgff3}
(\forall i \neq j) \quad \e^{-\eps} |P[i, j]| \leq |P'[i, j]| \leq \e^{\eps} |P[i, j]|.
\end{equation}
If for each $l$ $f_l^*$ denotes the optimal flow in~\eqref{e:traceapxgff1}, then
\begin{equation*}
\begin{split}
\TR\big(Q'[V, V]^{-1}\big)~&\leq~\sum_{l \in V}~\sum_{i \neq j } \frac{f_l^*(i, j)^2}{|P'[i, j]|} \quad (\textrm{by~\eqref{e:traceapxgff2}}) \\ 
~&\leq~\e^{\eps}~\sum_{l \in V}~\sum_{i \neq j} \frac{f_l^*(i, j)^2}{|P[i, j]|}~=~\e^{\eps} \TR\big(Q[V, V]^{-1}\big) \quad (\textrm{by~\eqref{e:traceapxgff3}~\textrm{and}~\eqref{e:traceapxgff1}}).
\end{split}
\end{equation*}
The proof of $\e^{-\eps} \TR\big(Q[V, V]^{-1}\big)~\leq~\TR\big(Q'[V, V]^{-1}\big)$ is identical.
\EPRF

Finally, we analyze the error of our approximation scheme. We first prove that the approximate messages are close to the ideal messages (Lemma~\ref{l:gffbotup}), and then show how to extract an approximately optimal solution from the approximate messages (Lemma~\ref{l:gfftopdown}).

\BLEM
\label{l:gffbotup}
Consider any $\eps > 0$ and any message $\alpha_{i \to j}$ of height $h$. For any $P_{ij}, Q_{ji}, S_{ij}, N_i$ for which $\alpha_{i\to j}\big(P_{ij}, Q_{ji}, S_{ij}, N_i\big)$ is finite and for any $\hat{Q}_{ji}\approxeq_{3^{2\kappa' h'}\eps} Q_{ji}$ there exists $\hat{P}_{ij}\approxeq_{3^{2\kappa'h}\eps} P_{ij}$ such that
\begin{equation}
\label{e:gffbotup}
\begin{split}
\hat{\alpha}_{i \to j}(\hat{P}_{ij}, \hat{Q}_{ji}, S_{ij}, N_i)~\leq~\exp\big(3^{2\kappa'\max\{h, h'\}+2\kappa'h}\eps\big)~\alpha_{i \to j}(P_{ij}, Q_{ji}, S_{ij}, N_i).
\end{split}
\end{equation}
\ELEM

\BPRF[of Lemma~\ref{l:gffbotup}:]
The proof is by induction on the height $h$ of the messages $\alpha_{i \to j},~\hat{\alpha}_{i \to j}$. The base case, $h = 1$, corresponding to messages sent from leaves, is easy to verify by comparing~\eqref{e:gpleaf} to~\eqref{e:agpleaf}. Let the minimum for the ideal message~\eqref{e:gpleaf} be achieved for $L_{ij} = L_{ij}^*$.  Then consider the objective for the approximate message~\eqref{e:agpleaf} with $\hat{L}_{ij} = L_{ij}^*$, $\hat{P}_{ij} = \mathrm{Round}_{\eps}(P_{ij})$, $\hat{S}_{ij} = S_{ij}$, $\hat{N}_i = N_i$, and $\hat{V}'_i = \Gamma_{ij} \setminus L_{ij}^* = V'_i$, where $V'_i$ is as defined in~\eqref{e:gpleaf}. Note that the objective in each of~\eqref{e:gpleaf},~\eqref{e:agpleaf} is the trace of the inverse of a submatrix obtained by applying the transformation $\mathrm{Obs}$ to sum of $\Lambda_{V_i}$ and $Q_{ji}, \hat{Q}_{ji}$ respectively. Since $\hat{Q}_{ji}\approxeq_{3^{2\kappa' h'}\eps} Q_{ji}$ (by hypothesis), we have by fact~\eqref{e:addapxgff} (for the sum) and fact~\eqref{e:obsapxgff} (for $\mathrm{Obs}$ and submatrix) that
\begin{equation*}
\mathrm{Obs}\big(\Lambda_{V_{i}} + \hat{Q}_{ji},~S_{ij} \cup L_{ij}^* \big) \approxeq_{3^{2\kappa'h'} \eps} \mathrm{Obs}\big(\Lambda_{V_i} + Q_{ji},~S_{ij} \cup L_{ij}^* \big).
\end{equation*}

Therefore,
\begin{equation*}
\frac{\hat{\alpha}_{i \to j}(\hat{P}_{ij}, \hat{Q}_{ji}, S_{ij}, N_i)}{\alpha_{i \to j}(P_{ij}, Q_{ji}, S_{ij}, N_i)}~\leq~\frac{\TR\bigg( \big(\hat{\Lambda}'_{V_i}[\hat{V}'_i, \hat{V}'_i] \big)^{-1} \bigg)}{\TR\bigg(\big( \Lambda'_{V_i}[V'_i, V'_i] \big)^{-1} \bigg)} 
~\leq~\exp\big(3^{2\kappa'h'} \eps \big),
\end{equation*}
where in the second step we used Lemma~\ref{l:traceapxgff} to account for taking the trace of inverse of precision matrices. Note that since  $\hat{P}_{ij} = \mathrm{Round}_{\eps}(P_{ij})$, we trivially have $\hat{P}_{ij} \approxeq_{3^{2\kappa'}\eps} P_{ij}$.

Hence it remains to verify the induction step for messages of height $h > 1$. Let $P_{ki}^*, Q_{ik}^*, S_{ik}^*, N_k^*, P_{li}^*,Q_{il}^*, S_{il}^*, N_l^*, L_{ij}^*$ be the optimal choice in~\eqref{e:compose'} for the given arguments $P_{ij}, Q_{ji}, S_{ij}$ and $N_i$. 

Now consider~\eqref{e:acompose'} with $\hat{S}_{ik}=S_{ik}^*$, $\hat{S}_{il}=S_{il}^*$, $\hat{N}_k=N_k^*$, $\hat{N}_l=N_l^*$ and $\hat{L}_{ij} = L_{ij}^*$. Let $\hat{P}_{ki}, \hat{P}_{li}$ be such that
\begin{equation}
\label{e:gffbotup6}
\hat{P}_{ki} \approxeq_{3^{2\kappa'(h-1)}\eps} P_{ki}^* \quad \textrm{and} \quad \hat{P}_{li} \approxeq_{3^{2\kappa'(h-1)}\eps} P_{li}^*.
\end{equation}

Further, let $\hat{Q}_{ik}, \hat{Q}_{il}$, and $\hat{P}_{ij}$ be given by~\eqref{e:acompose3} and~\eqref{e:acompose1} respectively.

Now $\hat{Q}_{ik}$ is obtained (see~\eqref{e:acompose3}) by successively applying transformations $\mathrm{Obs}$, $\mathrm{Marginal}_{V_i \setminus (\hat{S}_{ij} \cup \hat{L}_{ij}),~\Delta_{ik} \setminus \hat{S}_{ik}}$ and $\mathrm{Round}_{\eps}$ to the sum of $\Lambda_{V_i}, \hat{Q}_{ji}$ and $\hat{P}_{li}$. 
Hence by facts~\eqref{e:obsapxgff} (for $\mathrm{Obs}$),~\eqref{e:margapxgff} (for $\mathrm{Marginal}_{V_i \setminus (\hat{S}_{ij} \cup \hat{L}_{ij}),~\Delta_{ik} \setminus \hat{S}_{ik}}$),~\eqref{e:addapxgff} (for the sum), Observation~\ref{o:trianglegff} (for $\mathrm{Round}_{\eps}$) and~\eqref{e:gffbotup6}, we have that 
\begin{equation}
\label{e:gffbotup3}
\hat{Q}_{ik}\approxeq_{\eps'} Q_{ik}^*~\textrm{where}~\eps'~\leq~\eps + 3^{\kappa'} \max \big\{ 3^{2\kappa'(h-1)}\eps,~3^{2\kappa' h'}\eps \big\}~<~3^{2\kappa'\max\{h,~(h'+1)\}}\eps.
\end{equation}

Similarly we have $\hat{Q}_{il}\approxeq_{\eps'} Q_{il}^*$, and that (see~\eqref{e:acompose1}) 
\begin{equation}
\label{e:gffbotup4}
\hat{P}_{ij}\approxeq_{\eps''} P_{ij}^*~\textrm{where}~\eps''~\leq~\eps + 3^{\kappa'} 3^{2\kappa'(h-1)} \eps~<~3^{2\kappa'h} \eps.
\end{equation}

To complete the induction step we compare the corresponding terms on the r.h.s. of equations~\eqref{e:acompose'} and~\eqref{e:compose'}, which define how messages $\hat{\alpha}_{i \to j}$ and $\alpha_{i \to j}$ are composed respectively from $\hat{\alpha}_{k \to i},~\hat{\alpha}_{l \to i}$ and $\alpha_{k \to i},~\alpha_{l \to i}$.

First, by induction hypothesis there exists $\hat{P}_{ki}$ which satisfies~\eqref{e:gffbotup6}, i.e., $\hat{P}_{ki} \approxeq_{3^{2\kappa'(h-1)}\eps} P_{ki}^*$, and for which the messages $\hat{\alpha}_{k \to i}$ and $\alpha_{k \to i}$ (which are of height $h-1$) satisfy
\begin{equation}
\label{e:gffbotup1}
\begin{split}
\frac{\hat{\alpha}_{k \to i}(\hat{P}_{ki}, \hat{Q}_{ik}, \hat{S}_{ik}, \hat{N}_k)}{\alpha_{k \to i}(P_{ki}^*, Q_{ik}^*, S_{ik}^*, N_k^*)}~&\leq~\exp\bigg(3^{2\kappa'\max\{\max\{h,~(h'+1)\},~(h-1)\} +2\kappa'(h-1)} \eps\bigg) \\
~&\leq~\exp\big(3^{2\kappa'\max\{h,~h'\}+2\kappa'h} \eps \big),
\end{split}
\end{equation}
where we used the fact~\eqref{e:gffbotup3}  that $\hat{Q}_{ik}\approxeq_{\eps'} Q_{ik}^*$. 

Second, there exists $\hat{P}_{li}$ which satisfies~\eqref{e:gffbotup6}, i.e., $\hat{P}_{li} \approxeq_{3^{2\kappa'(h-1)}\eps} P_{li}^*$, and for which  similarly
\begin{equation}
\label{e:gffbotup1'}
\frac{\hat{\alpha}_{l \to i}(\hat{P}_{li}, \hat{Q}_{il}, \hat{S}_{il}, \hat{N}_l)}{\alpha_{l \to i}(P_{li}^*, Q_{il}^*, S_{il}^*, N_l^*)}~\leq~\exp\big(3^{2\kappa'\max\{h,~h'\}+2\kappa'h} \eps\big),
\end{equation}
 where we used the induction hypothesis and the fact that $\hat{Q}_{il}\approxeq_{\eps'} Q_{il}^*$.
 
Finally, it only remains to compare the trace terms in~\eqref{e:acompose'} and~\eqref{e:compose'}. Since by definition $\hat{P}_{ki} \approxeq_{3^{2\kappa'(h-1)}\eps} P_{ki}^*$ and $\hat{P}_{li} \approxeq_{3^{2\kappa'(h-1)}\eps} P_{li}^*$, and since by hypothesis $\hat{Q}_{ji}\approxeq_{3^{2\kappa' h'}\eps} Q_{ji}$, we have by facts~\eqref{e:obsapxgff} and~\eqref{e:addapxgff} that 
\begin{equation}
\label{e:gffbotup5}
{\hat{\Lambda}'}_{V_{i}} \big[\hat{V}'_i , \hat{V}'_i \big]~\approxeq_{\max\{3^{2\kappa'(h-1)},~3^{2\kappa'h'}\} \eps}~{\Lambda'}_{V_{i}} \big[ V'_i, V'_i \big]
\end{equation}
(where note that $\hat{V}'_i = V_i$ since $\hat{S}_{ij} = S_{ij}$ and $\hat{L}_{ij} = L_{ij}^*$). By Lemma~\ref{l:traceapxgff} and~\eqref{e:gffbotup5}, the ratio of the trace terms is bounded as
\begin{equation}
\label{e:gffbotup2}
\frac{\TR \bigg( \big({\hat{\Lambda}'}_{V_{i}} \big[\hat{V}'_i , \hat{V}'_i \big] \big)^{-1} \bigg)}
{\TR \bigg( \big({\Lambda'}_{V_{i}} \big[ V'_i, V'_i \big] \big)^{-1} \bigg)}~\leq~\exp\big(\max\{3^{2\kappa'(h-1)},~3^{2\kappa'h'}\} \eps\big).
\end{equation}
Combining~\eqref{e:gffbotup1},~\eqref{e:gffbotup1'} and~\eqref{e:gffbotup2} we obtain~\eqref{e:gffbotup}. Since we have already established in~\eqref{e:gffbotup4} that $\hat{P}_{ij}\approxeq_{3^{2\kappa'h} \eps} P_{ij}^*$, this completes the proof.
\EPRF

Lemma~\ref{l:gffbotup} establishes that the optimal value of approximate messages is not much larger than the optimal value of ideal messages (which corresponds to the optimal solution). The converse of Lemma~\ref{l:gffbotup} is stated in terms of the total error in a subtree given a set of observations in that subtree, which we defined in~\eqref{e:approxerr}. 

\BLEM
\label{l:gfftopdown}
Let $\hat{\alpha}_{i \to j}$ be a message of height $h$, and consider any $\hat{P}_{ij}, \hat{Q}_{ji}, \hat{S}_{ij}, \hat{N}_i$ 
for which $\hat{\alpha}_{i\to j}\big(\hat{P}_{ij}, \hat{Q}_{ij},  \hat{S}_{ij}, \hat{N}_i \big)$ is finite. Then the set of observations $S\subseteq V_{ij}$, obtained by solving the approximate dynamic program~\eqref{e:acompose'}-\eqref{e:acompose3} on subtree $T_{ij}$, is such that $S \cap \Delta_{ij} = \hat{S}_{ij}$, $|S| \leq \hat{N}_i$, and for any $Q_{ji}\approx_{3^{2\kappa'h'}\eps } \hat{Q}_{ji}$,
\begin{equation}
\begin{split}
\label{e:gfftopdown}
R_{ij}\big(Q_{ji}, S\big)~\leq~\exp\big(3^{2\kappa'\max\{h', h\} + 2\kappa'h}\eps\big) \hat{\alpha}_{i\to j}(\hat{P}_{ij}, \hat{Q}_{ji}, S_{ij}, N_i),
\end{split}
\end{equation}
and such that the distribution on ${\bf X}_{\Delta_{ij}}$ given observations in $S$ has precision $P_{ij}$ in $T_{ij}$, where $P_{ij}\approx_{3^{2\kappa'h}\eps}\hat{P}_{ij}$.
\ELEM

\BPRF[of Lemma~\ref{l:gfftopdown}:]
The proof is by induction on the height $h$ of $\hat{\alpha}_{i \to j}$. 

We first prove the base case of $h = 1$, i.e., when cluster $V_i$ is a leaf. Consider the optimal solution, say $\hat{L}_{ij}^*$, to the approximate program for leaves given by~\eqref{e:agpleaf}, and define $\hat{V}'_i = \Gamma_{ij} \setminus \hat{L}_{ij}^*$. Note that the optimal solution $S$ for subtree $T_{ij}$  in this case is simply given by $S = \hat{S}_{ij} \cup \hat{L}_{ij}^*$. 

Now, when cluster $V_i$ is a leaf the error function $R_{ij}$, as defined in~\eqref{e:approxerr}, happens to be
\begin{equation}
\label{e:gfftopdown1'}
\begin{split}
R_{ij}\big(Q_{ji}, S\big) &= \TR\bigg( \big(\Lambda'_{V_i}[\hat{V}'_i, \hat{V}'_i]\big)^{-1} \bigg), \quad \textrm{with} \\
\Lambda'_{V_i} &= \mathrm{Obs}\big(\Lambda_{V_i} + Q_{ji},~\hat{S}_{ij} \cup \hat{L}_{ij}^*\big).
\end{split}
\end{equation}

We prove the case $h = 1$ by taking the ratio of the r.h.s. of~\eqref{e:gfftopdown1'} to that of~\eqref{e:agpleaf}. First, notice that the matrices $\Lambda'_{V_i}$ in~\eqref{e:gfftopdown1'} and $\hat{\Lambda}'_{V_i}$ in~\eqref{e:agpleaf} are obtained by applying the transformation $\mathrm{Obs}$ to the sum of $\Lambda_{V_i}$ and respectively  $Q_{ji}$ and $\hat{Q}_{ji}$. Since $Q_{ji}\approx_{3^{2\kappa'h'}\eps} \hat{Q}_{ji}$ by hypothesis, we obtain by applying~\eqref{e:addapxgff} (for the sum) and~\eqref{e:obsapxgff} (for $\mathrm{Obs}$) that 
\begin{equation}
\label{e:gfftopdown11}
\Lambda'_{V_i}~\approxeq_{3^{2\kappa'h'}\eps}~\hat{\Lambda}'_{V_i}.
\end{equation}

Hence by Lemma~\ref{l:traceapxgff}, the ratio of the r.h.s. of~\eqref{e:gfftopdown1'} to the r.h.s. of~\eqref{e:agpleaf} is at most
\begin{equation}
\label{e:gfftopdown12}
\frac{\TR\bigg( \big(\Lambda'_{V_i}[\hat{V}'_i, \hat{V}'_i]\big)^{-1} \bigg)}{\TR\bigg( \big(\hat{\Lambda}'_{V_i}[\hat{V}'_i, \hat{V}'_i]\big)^{-1} \bigg)}
~\leq~\exp\big({3^{2\kappa'h'}\eps}\big),
\end{equation}
which is as desired. Notice that the constraint in~\eqref{e:agpleaf}, that the true precision $P_{ij}$ of ${\bf X}_{\Delta_{ij}}$ given observations $\hat{S}_{ij} \cup \hat{L}_{ij}$ satisfy $\hat{P}_{ij} = \mathrm{Round}_{\eps}(P_{ij})$, implies that $P_{ij}\approx_{3^{2\kappa'}\eps}\hat{P}_{ij}$ is trivially true. This completes the proof for $h = 1$.

Next consider a message $\hat{\alpha}_{i \to j}$ of height $h > 1$. Let the optimum in~\eqref{e:acompose'} for arguments $\hat{P}_{ij}, \hat{Q}_{ji}, S_{ij}, N_i$  be $\hat{P}_{ki}^*, \hat{Q}_{ik}^*, \hat{S}_{ik}^*, \hat{N}_k^*, \hat{P}_{li}^*, \hat{Q}_{il}^*, \hat{S}_{il}^*, \hat{N}_l^*, \hat{L}_{ij}^*$. The solution $S$  consists of
\begin{equation*}
\begin{split}
S = S_l \cup & S_k \cup \hat{L}_{ij}^* \cup \hat{S}_{ij} \quad \textrm{where} \\
S \cap \Delta_{ij} = \hat{S}_{ij}, &~S \cap \Gamma_{ij} = \hat{L}_{ij}^*, \quad \textrm{and} \\
S_l \subseteq V_{li}, &~S_l \cap \Delta_{il} = \hat{S}_{il}^*, \quad \textrm{and} \\
S_k \subseteq V_{ki}, &~S_k \cap \Delta_{ik} = \hat{S}_{ik}^*. 
\end{split}
\end{equation*}

Define $S_{ij} = \hat{S}_{ij}^*, S_{ik} = \hat{S}_{ik}^*, N_k = N_k^*, S_{il} = \hat{S}_{il}^*, N_l = N_l^*, L_{ij} = \hat{L}_{ij}^*$. Let $P_{ki}$ be the precision matrix of the distribution of ${\bf X}_{\Delta_{ik}}$ given observations $S_k$ in subtree $T_{ki}$, and let  $P_{li}$ be the precision matrix of the distribution of ${\bf X}_{\Delta_{il}}$ given observations $S_l$ in subtree $T_{li}$. Further, let $P_{ij}$, $Q_{ik}$ and $Q_{il}$ be as defined in respectively~\eqref{e:compose1} and~\eqref{e:compose3}.

First, notice that by the induction hypothesis for height $h-1$, the matrices $P_{ki}$, $P_{li}$ as defined above must satisfy
\begin{equation}
\label{e:gfftopdown2}
P_{ki} \approxeq_{3^{2\kappa'(h-1)}\eps} \hat{P}_{ki}^*, \quad \textrm{and}
\end{equation}

\begin{equation}
\label{e:gfftopdown3}
P_{li} \approxeq_{3^{2\kappa'(h-1)}\eps} \hat{P}_{li}^*.
\end{equation}

Now compare the definition of $P_{ij}$ in~\eqref{e:compose1} to that of $\hat{P}_{ij}$ in~\eqref{e:acompose1}, which involve an addition and application of $\mathrm{Obs}$ and $\mathrm{Marginal}$. We have by a combination of~\eqref{e:gfftopdown2},~\eqref{e:gfftopdown3}, and Lemma~\ref{l:gffops} (for the addition, $\mathrm{Obs}$ and $\mathrm{Marginal}$ operations) that 
\begin{equation}
\label{e:gfftopdown4}
P_{ij} \approxeq_{\eps + 3^{\kappa'} 3^{2\kappa'(h-1)}\eps} \hat{P}_{ij},~\textrm{which implies}~P_{ij} \approxeq_{3^{2\kappa' h}\eps} \hat{P}_{ij},
\end{equation}
as claimed.

Further comparing our definition of $Q_{ik}$ in~\eqref{e:compose3} to that of $\hat{Q}_{ik}^*$ in~\eqref{e:acompose3}, we have 
\begin{equation}
\label{e:gfftopdown7}
Q_{ik}~\approxeq_{\eps + 3^{\kappa'} \max\{3^{2\kappa'h'}, 3^{2\kappa'(h-1)}\} \eps} \hat{Q}_{ik}^*,~\textrm{which implies}~Q_{ik} \approxeq_{3^{2\kappa'\max\{h,~(h'+1)\}}\eps} \hat{Q}_{ik}^*,
\end{equation}
where we have used~\eqref{e:gfftopdown3} and the hypothesis $Q_{ji}\approx_{3^{2\kappa'h'}\eps } \hat{Q}_{ji}$ along with Lemma~\ref{l:gffops} to account for the addition, $\mathrm{Obs}$, $\mathrm{Marginal}$ and $\mathrm{Round}_{\eps}$ operations used in~\eqref{e:compose3} and~\eqref{e:acompose3}. Similarly, we have 
\begin{equation}
\label{e:gfftopdown8}
Q_{il} \approxeq_{3^{2\kappa'\max\{h,~(h'+1)\}}\eps} \hat{Q}_{il}^*.
\end{equation}

Next, note that the error function $R_{ij}$ (as we defined it in~\eqref{e:approxerr}) satisfies the recurrence
\begin{equation}
\label{e:gfftopdown1}
\begin{split}
& R_{ij} \big(\hat{Q}_{ji}, S \big) = R_{ki} \big(Q_{ik}, S_k \big) + R_{li} \big(Q_{il}, S_l \big) + \TR \bigg( \big(\Lambda'_{V_{i}} \big[\hat{V}'_i, \hat{V}'_i\big] \big)^{-1} \bigg),~\textrm{with} \\
& \hat{V}'_i = \Gamma_{ij} \setminus \hat{L}_{ij}^*,~\textrm{and}~\Lambda'_{V_i} = \mathrm{Obs}\bigg(\Lambda_{V_i} + P_{li} + P_{ki} + Q_{ji},~\hat{S}_{ij} \cup \hat{L}_{ij}^* \bigg).
\end{split}
\end{equation}

We are going to prove the induction step for $h > 1$ by comparing the corresponding terms in~\eqref{e:gfftopdown1} and~\eqref{e:acompose'}. By the induction hypothesis for message $\hat{\alpha}_{k \to i}$, which is of height $h-1$, we have 
\begin{equation}
\label{e:gfftopdown5}
\begin{split}
\frac{R_{ki}\big(Q_{ik}, S_k \big)}{\hat{\alpha}_{k \to i}\big(\hat{P}_{ki}, \hat{Q}_{ik}, \hat{S}_{ik}, \hat{N}_k \big)}
~&\leq~\exp\big(3^{2\kappa' \max\{ \max\{h, (h'+1)\}, h-1\} + 2\kappa'(h-1)}\eps\big) \\
~&\leq~\exp \big(3^{2\kappa'\max\{h', h\} + 2\kappa'h} \eps\big),
\end{split}
\end{equation}
where we used~\eqref{e:gfftopdown7}. Similarly, by the induction hypothesis for $\hat{\alpha}_{l \to i}$ (of height $h-1$) and by~\eqref{e:gfftopdown8} we get
\begin{equation}
\label{e:gfftopdown6}
\frac{R_{li}\big(Q_{il}, S_l \big)}{\hat{\alpha}_{l \to i}\big(\hat{P}_{li}, \hat{Q}_{il}, \hat{S}_{il}, \hat{N}_l \big)}~\leq~\exp \big(3^{2\kappa'\max\{h', h\} + 2\kappa'h} \eps\big).
\end{equation}

It only remains to compare the trace terms in~\eqref{e:gfftopdown1} and~\eqref{e:acompose'}. However note that by~\eqref{e:addapxgff} (for  addition) and~\eqref{e:obsapxgff} (for $\mathrm{Obs}$) combined with~\eqref{e:gfftopdown2},~\eqref{e:gfftopdown3} and our hypothesis that $Q_{ji} \approx_{3^{2\kappa'h'}\eps } \hat{Q}_{ji}$, we get
\begin{equation}
\label{e:gfftopdown9}
\Lambda'_{V_{i}}~\approxeq_{\max\{3^{2\kappa'(h-1)},~3^{2\kappa'h'}\} \eps} \hat{\Lambda}'_{V_{i}},~\textrm{which trivially implies}~\Lambda'_{V_{i}} \approxeq_{3^{2\kappa'\max\{h', h\} + 2\kappa'h}\eps} \hat{\Lambda}'_{V_{i}}.
\end{equation}

Finally by~\eqref{e:gfftopdown9},~\eqref{e:obsapxgff} (for $\mathrm{Obs}$) and Lemma~\ref{l:traceapxgff} (to account for the trace of the inverses), the ratio of the trace terms in~\eqref{e:gfftopdown1} and~\eqref{e:acompose'} is at most
\begin{equation}
\label{e:gfftopdown10}
\frac{\TR \bigg( \big(\Lambda'_{V_{i}} \big[\hat{V}'_i, \hat{V}'_i\big] \big)^{-1} \bigg)}{\TR \bigg( \big(\hat{\Lambda}'_{V_{i}} \big[\hat{V}'_i, \hat{V}'_i\big] \big)^{-1} \bigg)}~\leq~\exp\big(3^{2\kappa'\max\{h', h\} + 2\kappa'h} \eps\big).
\end{equation}

Combining~\eqref{e:gfftopdown5},~\eqref{e:gfftopdown6} and~\eqref{e:gfftopdown10} we get~\eqref{e:gfftopdown} as desired.
\EPRF

Lemmas~\ref{l:gffbotup} and~\ref{l:gfftopdown} show that the error introduced due to rounding off each element of the precision matrices can scale exponentially in the height of the tree-decomposition. In general, tree-decompositions can be unbalanced, i.e., have height $\Omega(n)$ (as in the case of, e.g., a simple $1$-D chain), implying that the error can scale exponentially in $n$. However, we will see in Section~\ref{sec:balancedtree} that there is an algorithm due to Bodlaender which always produces balanced tree-decompositions of height $O(\ln n)$, thereby allowing the error to be bounded as a polynomial in $n$ instead.

\subsubsection{Message Passing for GFFs Using Balanced Tree-decompositions}
\label{sec:balancedtree}

We first state Bodlaender's construction of a balanced binary tree-decomposition of the graph.
\BLEM
\label{l:shallowbinary}
(Theorem~4.2 of~\citet{Bod88} and Theorem~1.1 of~\citet{Bod96})
There exists an algorithm which, given any graph of tree-width $\kappa$ on $n$ nodes, constructs a tree-decomposition having width $\kappa' \leq 3\kappa + 2$, height at most $2\lceil \log_{5/4}(2n)\rceil$, size at most $20n$ and satisfying our requirements in Note~\ref{note:decompassume} (see page~\pageref{note:decompassume}). The running time of the algorithm is $n \cdot 2^{O(\kappa^3)}$.
\ELEM

The usual constructions (see, e.g.,~\citet{Bod96}) can produce tree-decompositions having height (and diameter) linear in $n$. Bodlaender's transformation~\citep{Bod88}, based on a result of~\citet{MR85}, takes a tree-decomposition as input and produces a wider but shallower decomposition (of  height logarithmic in $n$), which means the round-off error due to approximations in message passing would be smaller than that on the original tree-decomposition. Hence, by Lemma~\ref{l:shallowbinary}, we can assume that each cluster $V_i$ has size at most $\kappa'+1 \leq 3\kappa + 3$, that the number of nodes in the tree $T$ is $m \leq 20n$, and that the height of $T$ is at most $2\lceil \log_{5/4}(2n)\rceil$. This means that for GFFs, the element-wise rounding scheme for precision matrices yields an FPTAS.

\BTHM
\label{t:gfftwmain}
There is a dynamic programming algorithm which, for any $\kappa$, and any GFF on a graph on $n$ vertices with tree-width bounded by $\kappa$ does the following. For any $\frac{1}{2} > \eps' > 0$ and budget $b$,  it outputs a set $S_{b, \eps'}$ such that
$$\mathrm{err}\big(S_{b, \eps'}\big)~\leq (1 + \eps')~\min_{|S| \leq b}~\mathrm{err}\big(S\big).$$

The algorithm runs in time 
$$\bigg(n\ln\bigg( \frac{\max_{ \{i, j\} \in E} r_{ij}}{\min_{i \neq j} r_{ij}} \bigg)~\big/~\eps' \bigg)^{O(\kappa^3)} b^2.$$
\ETHM

We remark that the $\ln\frac{\max_{ \{i, j\} \in E} r_{ij}}{\min_{i \neq j} r_{ij}}$ term in the time complexity can be thought of as the number of bits required to describe the input GFF.

\BPRF[of Theorem~\ref{t:gfftwmain}:]
Consider the tree $T$ produced by the construction in Lemma~\ref{l:shallowbinary}. We set $\eps = \frac{\eps'}{4(2n)^{240\kappa}}$ and run the algorithm given by~\eqref{e:acompose'}-\eqref{e:acompose3} and~\eqref{e:agpleaf} on tree $T$, with the transformation $\mathrm{Round}_{\eps}$ and the $\eps$-nets $\big\{{\cal I}_{\eps}^{V \times V} \big\}$ as defined respectively in~\eqref{e:gffround} and in~\eqref{e:gffepsnet}. Recall that the cluster $V_m$ as per our assumption is empty, i.e., $V_m = \emptyset$, and is a leaf with neighbour (say) $V_i$ so that $\Delta_{im} = V_i$. Our output, $S_{b, \eps'}$, is simply the set of observations extracted from the approximate message $\hat{\alpha}_{i \to m}\big(\hat{P}_{im}^*, \hat{Q}_{mi}, \hat{S}_{im}^*, b\big)$ where $\hat{Q}_{mi}$ is an all-zeros matrix and where
$$\hat{P}_{im}^*,  \hat{S}_{im}^*~\define~\arg \underset{\hat{P}_{im},~\hat{S}_{im} \subseteq \Delta_{im}}{\min}~\hat{\alpha}_{i \to m}\big(\hat{P}_{im}, \hat{Q}_{mi}, \hat{S}_{im}, b\big).$$

Let the optimal solution be $S_b^*$ and let $P_{im}^*$ be the precision matrix of ${\bf X}_{\Delta_{im}} = {\bf X}_{V_i}$ given observations $S_b^*$. 
Now given that we set $\eps = \frac{\eps'}{4(2n)^{240\kappa}}$ and that height of $T$ is at most $2\lceil \log_{5/4}(2n)\rceil$, we have
\begin{equation*}
\begin{split}
\mathrm{err}\big(S_{b, \eps'}\big)~&=~\frac{1}{n} R_{im}\big(\hat{Q}_{mi}, S_{b, \eps'}\big) \quad (\textrm{by the definition of}~R_{im},~\textrm{see~\eqref{e:approxerr})} \\
~&\leq~\exp(\eps'/4)~\frac{1}{n}\hat{\alpha}_{i \to m}\big(\hat{P}_{im}^*, \hat{Q}_{mi}, \hat{S}_{im}^*, b\big)~~(\textrm{by Lemma~\ref{l:gfftopdown}}~\textrm{and our choice of}~\eps) \\
~&\leq~\exp(\eps'/4)~\frac{1}{n}\hat{\alpha}_{i \to m}\big(\hat{P}_{im}, \hat{Q}_{mi}, S_b^* \cap \Delta_{im}, b\big) \quad (\textrm{where}~\hat{P}_{im}~\textrm{is as defined in}  \\
~& \qquad \qquad \qquad \qquad \qquad \qquad \textrm{Lemma~\ref{l:gffbotup} for arguments}~P_{im}^*, \hat{Q}_{mi}, S_b^* \cap \Delta_{im}, b) \\
~&\leq~\exp(\eps'/4)\frac{1}{n} \bigg( \exp(\eps'/4)~\alpha_{i \to m}\big(P_{im}^*, \hat{Q}_{mi}, S_b^* \cap \Delta_{im}, b\big)\bigg) \\
~& \qquad \qquad \qquad \qquad (\textrm{by Lemma~\ref{l:gffbotup}}~\textrm{and our choice of}~\eps) \\
~&<~(1 + \eps')~\frac{1}{n}\alpha_{i \to m}\big(P_{im}^*, \hat{Q}_{mi}, S_b^* \cap \Delta_{im}, b\big)\bigg) \\
~&=~(1 + \eps')\mathrm{err}\big(S_b^*\big).
\end{split}
\end{equation*}

As for the running time, first recall that the time required to construct $T$ is $n 2^{O(\kappa^3)}$ by Lemma~~\ref{l:shallowbinary}. Next, note that all precision matrices produced by message passing have supports of size at most $\kappa' \times \kappa' \leq 16 \kappa^2$, and hence the size of each $\eps$-net used (by Observation~\eqref{o:gffepsnetsize}) is at most
$$\bigg(2 + \frac{1}{\eps}\ln\bigg( \frac{n^2}{2} \frac{\max_{ \{i, j\} \in E} r_{ij}^2}{\min_{i \neq j} r_{ij}^2} \bigg) \bigg)^{16\kappa^2},$$
which, given $\eps = \frac{\eps'}{4(2n)^{240\kappa}}$, is of the order 
$$\bigg(n\ln\bigg(n \frac{\max_{ \{i, j\} \in E} r_{ij}}{\min_{i \neq j} r_{ij}} \bigg)~\big/~\eps' \bigg)^{O(\kappa^3)}.$$

Hence using sparse representations for matrices, we can perform the message passing step~\eqref{e:acompose'} for each edge $\{i, j\}$ in $T$ in time 
$$\bigg(n\ln\bigg(n \frac{\max_{ \{i, j\} \in E} r_{ij}}{\min_{i \neq j} r_{ij}} \bigg)~\big/~\eps' \bigg)^{O(\kappa^3)} b^2,$$
which gives the claimed time complexity since by construction $T$ has $m \leq 20n$ edges. 
\EPRF


\subsection{Approximate message passing for Gaussian MRF}
\label{sec:GPtreewidthapprox}

In this section we describe a rounding scheme for general Gaussian MRFs on bounded tree-width graphs which is based on singular value decomposition and on the usual ordering $\preceq$ of positive semidefinite matrices. The element-wise rounding of precision matrices for GFFs (see Section~\ref{sec:gffround}) does not work for general Gaussian MRFs whose precision matrices can be arbitrary positive definite matrices instead of simple graph Laplacians. Note that we will be using the same approximate message passing scheme given by~\eqref{e:acompose'}-\eqref{e:acompose3} and~\eqref{e:agpleaf} as for GFFs, except that the transformation $\mathrm{Round}_{\eps}$ and the $\eps$-nets for precision matrices will be different. 

Our analysis of the error is going to be on similar lines as that for GFFs in Section~\ref{sec:gffround}. 
However, the size of the $\eps$-nets is going to depend polynomially on the condition number of the precision matrix $\Lambda$. As a consequence, our proposed algorithm does not yield a FPTAS for Gaussian MRFs with arbitrary precision matrices unless we impose restrictions on the condition number. In contrast, the size of the $\eps$-nets for GFFs (based on element-wise rounding of the precision matrices)  scales polylogarithmically in the size of the description of the input.

Before we describe the rounding scheme for general Gaussian MRFs, we point out that for the special case of trees ($\kappa = 1$), any Gaussian MRF is equivalent to a GFF---see Lemma~\ref{l:gffgp} below. Hence we get a FPTAS for general Gaussian MRF on trees as a corollary of Theorem~\ref{t:gfftwmain}.

\BLEM
\label{l:gffgp}
Consider any Gaussian MRF ${\bf X} = (X_1, X_2, \dots, X_n)$ on a tree $T$ with $n$ vertices such that no 2 variables are independent, i.e., for each pair $X_i, X_j$, $\ee[X_i X_j] \neq 0$. Then there exists a vector ${\bf w} \in \R^n$ and a GFF $(Y_1, Y_2, \dots,  Y_{n+k})$ on a tree $T'$ where $k \leq n$ (i.e., $T'$ has at most $2n$ vertices) such that $(w_1X_1, w_2X_2, \dots, w_nX_n)$ has the same distribution as that of $(Y_1, Y_2, \dots, Y_n)$ given that $(Y_{n+1}, Y_{n+2}, \dots, Y_{n+k})$ are observed.
\ELEM

\BPRF[of Lemma~\ref{l:gffgp}:]
 As usual, $\Lambda$ denotes the precision matrix for ${\bf X}$. Assume w.l.o.g. that $T$ is oriented such that vertex $j$ is a child of vertex $i$ only if $i < j$. We set the values $w_1, w_2, \dots, w_n$ in the following sequence. We first set $w_1$ such that for each child $j$ of $1$, $\big|\frac{\Lambda[j, 1]}{w_1}\big| < \Lambda[j, j]$. Consider any vertex $i \geq 1$ such that $w_i$ has been set, and let $i$ have $l$ children, $j_1, j_2, \dots, j_l > i$. We set $w_{j_1}, w_{j_2}, \dots, w_{j_l}$ (i.e., we scale each $X_{j_t}$ by $w_{j_t}$) to be such that:
\begin{itemize}
\item for each $t \leq l$,  $\frac{\Lambda[i, j_t]}{w_{i} w_{j_t}} \leq 0$, i.e., each off-diagonal entry in row $i$ is non-positive after scaling, and 
\item row $i$ is diagonally dominant after scaling, i.e.
$$\sum_{j} \frac{\Lambda[i, j]}{w_i w_j}~=~\frac{\Lambda[i, i]}{w_i^2} + \bigg(\sum_{j < i} \frac{\Lambda[i, j]}{w_i w_j}\bigg) + \bigg(\sum_{t = 1}^{l} \frac{\Lambda[i, j_t]}{w_{i} w_{j_t}} \bigg)~\geq~0,$$
and moreover
\item for each $t = 1,2, \dots, l$, $\big|\frac{\Lambda[j_t, i]}{w_{i} w_{j_t}}\big| < \Lambda[j_t, j_t]$.
\end{itemize}
We stop after having assigned a value to each $w_i$. This means that the variables $(w_1 X_1, w_2 X_2, \dots, w_n X_n)$ have a diagonally dominant precision matrix, say $\Lambda'$, with non-positive off-diagonal entries. 

Now, for each row $i$ of $\Lambda'$ which is strictly diagonally dominant (i.e. the row sum is positive), we add a new node to the tree $T$ and connect it  to node $i$ by an edge. The tree $T'$ thus obtained has $n + k$ nodes, where $k \leq n$.
Note that $\Lambda'$ is a principal submatrix, $\Lambda''[\{1, 2, \dots, n\}, \{1, 2, \dots, n\}]$, of a Laplacian $\Lambda''$ of size $n+k \times n+k$ on the tree $T'$, where the rows and columns of $\Lambda''$ indexed by $n+1,n+2, \dots, n+k$  are defined as
$$(\forall~n+1 \leq j \leq n+k) (\forall~i \neq j) \quad \Lambda''[j, i]~= \left\{ 
\begin{array}{cc}
-\sum_{l \leq n}\Lambda'[l, i] & \textrm{if}~i~\textrm{is the only node to} \\
~& \textrm{which}~j~\textrm{has an edge in}~T' \\
0 & \textrm{otherwise.}
\end{array} \right.
$$ 

We complete our proof by letting $(Y_1, Y_2, \dots,  Y_{n+k})$ be the GFF defined by the Laplacian $\Lambda''$ on $T'$, and observing that the principal submatrix $\Lambda''[\{1, 2, \dots, n\}, \{1, 2, \dots, n\}]$ is the precision matrix of $(Y_1, Y_2, \dots, Y_n)$ given that $Y_{n+1}, \dots, Y_{n+k}$ are observed. Note that we need the assumption that no 2 variables in the original Gaussian MRF are independent only to make sure that the tree $T'$ is connected---this is a requirement in our definition of GFFs. 
\EPRF

Before we describe our rounding idea for larger tree-widths, i.e. $\kappa \geq 1$, we make the following observation about the eigenvalues of all precision matrices obtained while running the ideal message passing algorithm (see~\eqref{e:compose'}-\eqref{e:compose3} and~\eqref{e:gpleaf}). We need this property since our construction of the $\eps$-nets for precision matrices requires the eigenvalues of the latter to be in a bounded range.

\BOBS
\label{o:eigenmessage}
Consider any edge $\{i, j\}$ in the given tree-decomposition. Consider precision matrices $P, Q \in {\cal X}_+^{n \times n}$ such that there exists some set $S \subseteq \Delta_{ij}$  and some integer $N$  for which $\alpha_{i \to j}[P, Q, S, N] < +\infty$. Then $P$ and $Q$ both have support $(\Delta_{ij} \setminus S) \times (\Delta_{ij} \setminus S)$ and rank $|\Delta_{ij} \setminus S|$. Moreover, 
$\frac{\lambda_{min}(\Lambda)}{m} \leq \lambda_{min}(P) \leq \lambda_{max}(P) \leq \lambda_{max}(\Lambda)$ and $\frac{\lambda_{min}(\Lambda)}{m} \leq \lambda_{min}(Q) \leq \lambda_{max}(Q) \leq \lambda_{max}(\Lambda)$.
\EOBS

Observation~\ref{o:eigenmessage} holds because of the following 2 reasons. First, the way we split the overall joint precision matrix $\Lambda$ into factors for each cluster (see~\eqref{e:gpdecomp1} in Lemma~\ref{l:gpdecomp}) ensures that each factor $\Lambda_{V_i}$ has rank $|V_i|$ and smallest non-zero eigenvalue at least $\frac{\lambda_{min}(\Lambda)}{m}$. Second, in the message composition steps~\eqref{e:compose1} and~\eqref{e:compose3} of our ideal algorithm, we use the transformations $\mathrm{Obs}$ and $\mathrm{Marginal}$ which, by Lemma~\ref{l:eigenpreserve}, do not make the smallest non-zero eigenvalues any smaller.

\BDEF 
Given 2 matrices $Q, Q' \in {\cal X}_{+}^{n \times n}$, and any $\eps \geq 0$, we will say that $Q' \approx_{\eps} Q$ iff $\e^{-\eps} Q~\preceq~Q'~ \preceq~\e^{\eps} Q$. 
\EDEF

As with the element-wise rounding for GFFs, we will need the following facts about $\approx_{\eps}$ in order to analyze how round-off error accumulates during message passing. 

\BOBS
\label{o:traceapprox}
For any set $V$, if $Q, Q' \in {\cal X}_{+}^{V \times V}$ are such that $Q'~\approx_{\eps}~Q$, and $Q[V, V]$ has full rank (i.e., $|V|$), then 
$$\e^{-\eps}~\TR\big(Q[V, V]^{-1}\big)~\leq~\TR\big(Q'[V, V]^{-1}\big)~\leq~\e^{\eps}~\TR\big(Q[V, V]^{-1}\big).$$
\EOBS

\BOBS
\label{o:triangle}
If $Q, Q_1, Q_2 \in {\cal X}_{+}^{n \times n}$ and $\eps_1, \eps_2 \geq 0$ are such that $Q_1 \approx_{\eps_1} Q$ and $Q_2 \approx_{\eps_2} Q$, then $Q_1 \approx_{\eps_1 + \eps_2} Q_2$. 
\EOBS

Observation~\ref{o:traceapprox} follows from the fact that if $Q \approxeq_{\eps} Q'$ then $Q[V, V] \approx_{\eps} Q'[V, V]$ as well, and hence by Corollary~7.7.4 of~\citet{HJ85}, the set of eigenvalues of $Q[V, V]$ and $Q'[V, V]$ are within a factor $\e^{\pm \eps}$ of each other. Observation~\ref{o:triangle} is a consequence of transitivity of positive semidefinite ordering.

\BLEM
\label{l:gpops}
Consider any $\eps \geq 0$. Then for any set $V$ and any $Q, Q' \in {\cal X}_{+}^{V \times V}$ such that $Q'~\approx_{\eps}~Q$, we have 
\begin{equation}
\label{e:obsapx}
(\forall O \subseteq V) \quad \mathrm{Obs}\big(Q' ,~O\big)~\approx_{\eps}~\mathrm{Obs}\big(Q,~O\big), \quad \textrm{and}
\end{equation}
\begin{equation}
\label{e:margapx}
(\forall \Delta \subseteq V) \quad \mathrm{Marginal}_{V,\Delta}\big(Q'\big)~\approx_{\eps}~\mathrm{Marginal}_{V,\Delta}\big(Q\big).
\end{equation}
Further, for any $Q_1, Q_2$ and $Q'_1, Q'_2$ in ${\cal X}_{+}^{n \times n}$ such that $Q'_1 \approx_{\eps} Q_1$ and $Q'_2 \approx_{\eps} Q_2$, we have
\begin{equation}
\label{e:addapx}
Q'_1 + Q'_2~\approx_{\eps}~Q_1 + Q_2.
\end{equation}
\ELEM

\BPRF[of Lemma~\ref{l:gpops}:]
Equation~\eqref{e:obsapx} follows from the fact that $0 \preceq A$ implies $0 \preceq A[V, V]$. Equation~~\eqref{e:margapx} follows from the fact that if $A\preceq B$ then $B^{-1}\preceq A^{-1}$ (see, e.g.,~\citet{HJ85}, Corollary 7.7,4). Finally, equation~\eqref{e:addapx} follows from the fact that $0 \preceq A $ and $0 \preceq B$ imply $0 \preceq A+B$.
\EPRF

Next we describe how to construct, for any set $V$ and any $\frac{1}{4} \geq \eps > 0$, an $\eps$-net for ${\cal X}_+^{V \times V}$, denoted by ${\cal I}_{\eps}^{V \times V}$. We will be rounding the precision matrices obtained during message passing by first performing a singular value decomposition or SVD (see, e.g., Chapter~7.3 of~\citet{HJ85}), and then separately rounding the orthogonal matrix and diagonal matrix (containing the eigenvalues) thus obtained to $\eps$-nets for orthogonal matrices and for diagonal matrices respectively.  

The first ingredient in our construction is an $\eps$-net ${\cal R}_{\eps}^{k \times k}$ for all $k \times k$ orthogonal matrices.

\BLEM
\label{l:epsortho}
For any $k$ and $\frac{1}{k(4\sqrt{2} + 4)} \geq \eps > 0$ there exists a set ${\cal R}_{\eps}^{k \times k}$ of orthogonal matrices such that 
$$|{\cal R}_{\eps}^{k \times k}|~\leq~\big( 2k\e^{\eps k / 2}\big(2/\eps\big)^{k-1} \big)^k,$$
and such that for each $k \times k$ orthogonal matrix $U$, there exists a $U'' \in {\cal R}_{\eps}^{k \times k}$ which satisfies
\begin{equation}
\label{e:epsortho}
(\forall i) \quad (U''[:,i])^t U[:,i] > 1 - 12.5k\eps~~\textrm{and}~~(\forall j \neq i)~(U''[:,i])^t U[:,j] < 10(\sqrt{2} + 1)\sqrt{k\eps}.
\end{equation}
\ELEM

Intuitively, Lemma~\ref{l:epsortho} means that columns of the approximation $U''$ of an orthogonal matrix $U$ can be obtained from the columns of $U$ by small rotations. We defer the proof of Lemma~\ref{l:epsortho} until later. The second ingredient in the construction is an $\eps$-net for eigenvalues, which are between $\frac{1}{m}\lambda_{min}(\Lambda)$ and $\lambda_{max}(\Lambda)$, defined as
\begin{equation}
\label{e:epseigen}
\begin{split}
{\cal L}_{\eps}~&=~\bigg\{\frac{\lambda_{min}(\Lambda)}{m},~ \e^{\eps}\frac{\lambda_{min}(\Lambda)}{m},~\e^{2\eps}\frac{\lambda_{min}(\Lambda)}{m}, \dots,~\e^{K\eps}\frac{\lambda_{min}(\Lambda)}{m}\bigg\},~\textrm{where} \\
K~&=~\bigg\lfloor \frac{1}{\eps}\ln\bigg(m\frac{\lambda_{max}(\Lambda)}{\lambda_{min}(\Lambda)} \bigg) \bigg\rfloor.
\end{split}
\end{equation}

We are now ready to define our $\eps$-net for precision matrices.

\BDEF
\label{def:gpepsnet}
For any set $V$, we define ${\cal I}_{\eps}^{V \times V} \subseteq {\cal X}_+^{V \times V}$ to be the collection of all matrices $A$ with support $V \times V$ such that $A[V, V] = R~D~R^t$ where  $R \in {\cal R}_{\eps_1}^{|V| \times |V|}$ and where $D$ is a $|V| \times |V|$ diagonal matrix whose diagonal entries belong to the set ${\cal L}_{\eps_2}$, with 
\begin{equation}
\label{e:orthochoice}
\eps_1 = \bigg(\frac{\lambda_{min}(\Lambda)}{m\lambda_{max}(\Lambda)} \bigg)^2 \frac{\eps^2}{10^4 |V|^3} \quad \textrm{and} \quad \eps_2 = \eps/2.
\end{equation}
\EDEF

Clearly, 
\begin{equation}
\label{e:sizeeps}
\big|{\cal I}_{\eps}^{V \times V}\big|~=~|{\cal L}_{\eps_2} \big|~\big| {\cal R}_{\eps_1}^{|V| \times |V|}\big|~=~
O\bigg( \bigg(\frac{m^2 |V|^3}{\eps^2}\bigg(\frac{\lambda_{max}(\Lambda)}{\lambda_{min}(\Lambda)} \bigg)^2 \bigg)^{|V|^2} \bigg).
\end{equation}

We have chosen the  parameters $\eps_1, \eps_2$ to be sufficiently small such that for any precision matrix $P \in {\cal X}_+^{V \times V}$ of rank $|V|$ with eigenvalues in the range $\big[\frac{\lambda_{min}(\Lambda)}{m},~\lambda_{max}(\Lambda)\big]$, there is a matrix $P'$ in ${\cal I}_{\eps}^{V \times V}$ which is $\eps$-close to $P$, i.e., $P \approx_{\eps} P'$. However, by Observation~\ref{o:eigenmessage}, all precision matrices $P$ produced while message passing have eigenvalues precisely in the above range. Hence ${\cal I}_{\eps}^{V \times V}$ is indeed a sufficiently fine $\eps$-nets for our approximation purposes. 
We next formally define the transformation $\mathrm{Round}_{\eps}$ which, as one would expect, works by first performing a SVD of the input precision matrix and then rounding the resulting orthogonal matrix and the diagonal matrix.

 \BLEM
 \label{l:epsmessage}
 For any $\eps \leq \frac{1}{4}$ there is a matrix transformation $\mathrm{Round}_{\eps}$ with the following property. For any set $V$ and for any $P \in {\cal X}_+^{V \times V}$ of rank $|V|$ satisfying $\frac{\lambda_{min}(\Lambda)}{m} \leq \lambda_{min}(P) \leq \lambda_{max}(P) \leq \lambda_{max}(\Lambda)$, we have that $\mathrm{Round}_{\eps}\big(P\big)~\in~{\cal I}_{\eps}^{V \times V}$ and that $\mathrm{Round}_{\eps}\big(P\big) \approx_{\eps}  P$. Moreover, $\mathrm{Round}_{\eps}\big(P\big)$ can be computed from $P$ in time $O(|V|^3)$.
 \ELEM

\BPRF[of Lemma~\ref{l:epsmessage}:]
We compute $P' = \mathrm{Round}_{\eps}\big(P\big),~P' \in {\cal X}_+^{V \times V}$ as follows. First compute the SVD  of $P[V, V] = U D U^t$, where $D$ is a diagonal matrix and $U$ is a $|V| \times |V|$ orthogonal matrix. Note that the diagonal entries of $D$, namely the eigenvalues of $P$, are in the range $\big[\lambda_{min}(\Lambda) / m,~\lambda_{max}(\Lambda)\big]$ as per our assumption in the lemma statement. Second, compute diagonal matrix $D'$ by rounding-off each entry of $D$ to the nearest element in ${\cal L}_{\eps/2}$ , i.e., for each $i$, $D'[i,i] = \arg \underset{r \in {\cal L}_{\eps/2}}{\min}~\big|r - D[i,i] \big|$. Next approximate $U$ using an orthogonal matrix $U' \in {\cal R}_{\eps_1}^{|V| \times |V|}$, where $\eps_1$ is as defined in~\eqref{e:orthochoice}, using the construction in Lemma~\ref{l:epsortho}. Finally define $P'[V, V] =  U' D' U'^t$. We can clearly perform all the steps of the computation in $O(|V|^3)$ time. \\

To prove $\mathrm{Round}_{\eps}\big(P\big) \approx_{\eps}  P$, we show that $U' D' U'^t = P'[V, V] ~\preceq~\e^{\eps} P[V, V] = \e^{\eps} U D U^t$. The proof of the other inequality, i.e., $\e^{-\eps} P[V, V] \preceq P'[V, V]$, is identical and omitted. To prove $U' D' U'^t \preceq \e^{\eps} U D U^t$, it is sufficient to prove that 
$$\e^{\eps}D - U^{-1}U'D'U'^t(U^t)^{-1} \succeq 0.$$

However the l.h.s. of the above can be simplified as
\begin{equation}
\label{e:epsmessage1}
\begin{split}
\e^{\eps}D - & U^{-1}U'D'U'^t(U^t)^{-1} \\
~&=~\e^{\eps}D - (U^tU') D'(U^tU')^{t} \\
~&=~\big(\e^{\eps} D - R D' R^t \big) \qquad (\textrm{substituting}~R = U^tU') \\
~&=~\big(\e^{\eps} D - D' \big) + \big(D' - RD'R^t \big) \\
~&\succeq~(\e^{\eps/2}D' -  D') + \big(D' - RD'R^t \big) \quad \bigg(\textrm{since}~(\forall i)~\frac{D'[i,i]}{D[i,i]} \leq \e^{\frac{\eps}{2}}\bigg) \\
~&\succeq~\frac{\eps}{2}  D'  + \big(D'R'^t + R'D' + R'R'^t \big) \quad \big(\textrm{substituting}~R' = R - I_{|V|}\big) \\
~&\succeq~\frac{\eps}{2}  D' + D'R'^t + R'D'. \\
\end{split}
\end{equation}

To see that $\frac{\eps}{2}  D' + D'R'^t + R'D' \succeq 0$, note that for any vector ${\bf z}$ of size $|V|$, 
\begin{equation}
\label{e:epsmessage2}
\begin{split}
{\bf z}^t & \bigg(\frac{\eps}{2}  D' + D'R'^t + R'D'\bigg) {\bf z} \\
~&\geq~\sum_{i}\bigg(\frac{\eps}{2} - \frac{25\eps_1 |V|}{2}\bigg)D'[i,i] z_i^2 \\
~& \qquad \qquad - 10(\sqrt{2} + 1)\sqrt{\eps_1 k}\sum_{i \neq j} (D'[i,i] + D'[j,j]) |z_i| |z_j| \\
~&\geq~\frac{\eps}{4|V|}\sum_{i \neq j}\bigg(D'[i,i] z_i^2 + D'[j,j] z_j^2 - 2 \min \big\{ D'[i,i], D'[j,j] \big\} |z_i| |z_j| \bigg) \\
~&\geq~0,
\end{split}
\end{equation} 
where in the first step we applied Lemma~\ref{l:epsortho},~\eqref{e:epsortho} to matrix $R' = R - I_{|V|} = U^tU' - I_{|V|}$, and in the second step we substituted the value of $\eps_1$ given by~\eqref{e:orthochoice}.
Hence $P'[V, V] ~\preceq~\e^{\eps} P[V, V]$.
\EPRF

Finally we give a proof of the construction of our $\eps$-net for orthogonal matrices, Lemma~\ref{l:epsortho}, which requires an $\eps$-net for unit vectors.

\BLEM
\label{l:epssphere}
For any $k$ and $\eps > 0$, there exists an $\eps$-net ${\cal U}_{\eps}^k$ for $k$-dimensional unit vectors where 
$$|{\cal U}_{\eps}^k|~<~2k\e^{\eps k / 2}\big(2/\eps\big)^{k-1},$$ 
and where 
\begin{equation}
\label{e:epssphere}
(\forall {\bf z}~\textrm{s.t.}~ \|{\bf z}\| = 1)~(\exists~{\bf z}' \in {\cal U}_{\eps}^k) \quad \| {\bf z}' - {\bf z} \|~\leq~\eps~~\textrm{and}~~{\bf z}^t {\bf z}'~\geq~1-\frac{\eps^2}{2}.
\end{equation}
\ELEM
\BPRF[sketch for Lemma~\ref{l:epssphere}:]
See, e.g., Lemma~5.5 of~\citet{DGL96}.
\EPRF

The  following lemma will also be useful in the proof of Lemma~\ref{l:epsortho}.
\BLEM
\label{l:orthoutil}
For any $1 \geq \eps > 0$, if ${\bf u}_1, {\bf v}_2,  {\bf z}_1, {\bf z}_2$ are unit vectors such that ${\bf u}_1^t {\bf u}_2 = 0$, $\| {\bf u}_1 - {\bf z}_1 \| \leq \eps$ and $\| {\bf u}_2 - {\bf z}_2 \| \leq \eps$, then $| {\bf z}_1^t {\bf z}_2| \leq (2\sqrt{2} + 2)\eps$.
\ELEM

\BPRF[of Lemma~\ref{l:orthoutil}:]
By triangle inequality,
\begin{equation}
\label{e:orthtutil1}
\begin{split}
\| {\bf z}_1 - {\bf z}_2 \|~&\leq~\| {\bf u}_1 - {\bf u}_2\| + \| {\bf u}_1 - {\bf z}_1 \| + \| {\bf u}_2 - {\bf z}_2\|, \\
~&\leq~\big(\sqrt{2}+2\eps\big), 
\end{split}
\end{equation}
where in the second inequality follows from the fact that $\| {\bf u}_1 - {\bf u}_2\| = \sqrt{2},~\| {\bf u}_1 - {\bf z}_1 \| \leq \eps$, and $\| {\bf u}_2 - {\bf z}_2 \| \leq \eps$.
A similar application of the triangle inequality gives
\begin{equation}
\label{e:orthtutil2}
\| {\bf z}_1 - {\bf z}_2 \|~\geq~\big(\sqrt{2} - 2\eps\big), 
\end{equation}

After taking squares of both sides of~\eqref{e:orthtutil1} and~\eqref{e:orthtutil2}, we get $\big| {\bf z}_1^t {\bf z}_2 \big|~\leq~2\sqrt{2}\eps + 2\eps^2~\leq~(2\sqrt{2} + 2)\eps$.
\EPRF

\BPRF[of Lemma~\ref{l:epsortho}:] 
${\cal R}_{\eps}^{k \times k}$ consists of all (orthogonal) matrices obtained by applying Gram-Schmidt orthonormalization (see, e.g., Chapter 0.6 of~\citet{HJ85}) to any $k \times k$ matrix with columns chosen from ${\cal U}_{\eps}^k$. Clearly 
$$|{\cal R}_{\eps}^{k \times k}|~\leq~\big| {\cal U}_{\eps}^k \big|^k~\leq~\big( 2k\e^{\eps k / 2}\big(2/\eps\big)^{k-1} \big)^k.$$
Now consider any $k \times k$ orthogonal matrix $U$ with columns $U[:,1], U[:,2], \dots, U[:,k]$. The matrix $U''$ in~\eqref{e:epsortho} can be computed from $U$ in time $O\big(k^3\big)$ by first rounding-off each column $U[:,i]$ to vector ${\bf u}'_i \in {\cal U}_{\eps}^k$, and then applying Gram-Schmidt orthonormalization to obtain columns $U''[:,1], U''[:,2], \dots, U''[:,k]$.  For the rounding-off step, we have by Lemmas~\ref{l:epssphere} and~\ref{l:orthoutil} that
\begin{equation}
\label{e:epsortho1}
(\forall i) \quad \| {\bf u}'_i - U[:,i] \| \leq \eps~\textrm{and}~(\forall j \neq i)~|({\bf u}'_i)^t {\bf u}'_j | \leq (2\sqrt{2} + 2)\eps.
\end{equation}

Next we analyze the Gram-Schmidt orthonormalization process. Consider for any $i$ the projection of ${\bf u}'_{i}$ onto the span of ${\bf u}'_1, {\bf u}'_2, \dots, {\bf u}'_{i-1}$, given by
\begin{equation}
\label{e:epsortho2}
{\bf u}'_{i \parallel} = \sum_{j < i} c_i {\bf u}'_j.
\end{equation}

First 
\begin{equation*}
\begin{split}
\| {\bf u}'_{i \parallel} \|^2 &= \sum_{j < i} c_j^2 \| {\bf u}'_j \|^2~+~\sum_{\substack{j_1 \neq j_2 \\ j_1, j_2 \neq i}} 2 c_{j_1} c_{j_2} ( {\bf u}'_{j_1})^t {\bf u}'_{j_2} \\
~&\geq~\sum_{j < i} c_j^2~+~(2\sqrt{2} + 2)\eps \sum_{\substack{j_1 \neq j_2 \\ j_1, j_2 \neq i}} 2 c_{j_1} c_{j_2} \quad (\textrm{using}~\eqref{e:epsortho1}) \\
~&\geq~\sum_{j < i} \big(1 - (i-2)(2\sqrt{2} + 2)\eps \big) c_j^2~+~(2\sqrt{2} + 2)\eps\sum_{\substack{j_1 \neq j_2 \\ j_1, j_2 \neq i}} \big(c_{j_1} - c_{j_2}\big)^2 \\
~&\geq~\big(1 - k(2\sqrt{2} + 2)\eps \big)\sum_{j < i}  c_j^2,
\end{split}
\end{equation*}
which, along with the hypothesis that $\frac{1}{k(4\sqrt{2} + 4)} \geq \eps$, implies

\begin{equation}
\label{e:epsortho3}
\sum_{j < i} c_j^2~\leq~\frac{\| {\bf u}'_{i \parallel} \|^2}{\big(1 - k(2\sqrt{2} + 2)\eps \big)}~\leq~\frac{1}{\big(1 - k(2\sqrt{2} + 2)\eps \big)}~\leq~2.
\end{equation}

Second, by~\eqref{e:epsortho1} we get
\begin{equation}
\label{e:epsortho4}
\begin{split}
\big|\big({\bf u}'_{i \parallel}\big)^t {\bf u}'_i\big|~&=~\big|\sum_{j < i} c_i ({\bf u}'_j)^t {\bf u}'_i\big| \\
~&\leq~(2\sqrt{2} + 2)\eps\sum_{j < i}|c_j| \\
~&\leq~(2\sqrt{2} + 2)\eps\sqrt{2(i-1)} \\
~&\leq~2(2+\sqrt{2})\sqrt{k}\eps,
\end{split}
\end{equation}
where in the penultimate step we used the fact that given~\eqref{e:epsortho3}, $\sum_{j < i}|c_j|$ can be at most $\sqrt{2(i-1)}$. The column $U''[:,i]$ produced by Gram-Schmidt is $U''[:,i] = \frac{ {\bf u}'_i - {\bf u}'_{i \parallel}}{\| {\bf u}'_i - {\bf u}'_{i \parallel} \|}$. Hence using~\eqref{e:epsortho4} we get $ (U''[:,i])^t {\bf u}'_i \geq ({\bf u}'_i - {\bf  u}'_{i \parallel})^t {\bf u}'_i \geq 1 - 2(2+\sqrt{2})\sqrt{k}\eps$, which in turn implies that
\begin{equation}
\label{e:epsortho6}
\| U''[:,i] - {\bf u}'_i \|~\leq~2\sqrt{(2+\sqrt{2})} \sqrt{k\eps}.
\end{equation}

Applying the triangle inequality along with~\eqref{e:epsortho1},~\eqref{e:epsortho6} we obtain
\begin{equation}
\label{e:epsortho5}
\|U''[:,i] - U[:,i]\|~\leq~\| U''[:,i] - {\bf u}'_i \| + \| {\bf u}'_i - U[:,i] \|~\leq~2\sqrt{(2+\sqrt{2})}\sqrt{k\eps} + \eps~<5\sqrt{k\eps}.
\end{equation}

It follows from~\eqref{e:epsortho5} that $(U''[:,i])^t U[:,i] > 1 - 12.5k\eps$.
Combining~Lemma~\ref{l:orthoutil} with~\eqref{e:epsortho5} gives
$$(\forall i \neq~j) \quad (U''[:,i])^t U[:,j]~<~10(\sqrt{2} + 1)\sqrt{k\eps}.$$
\EPRF

Having described how to construct $\eps$-nets, we are finally ready to analyze how the error due to rounding accumulates during message passing. Lemma~\ref{l:gpbotup} states that the approximate messages are not much bigger than the ideal messages, and Lemma~\ref{l:gfftopdown} shows thats the true error of the approximately optimal set of observations (extracted from the approximate messages) is not much bigger than the approximate error (given by approximate messages). 

\BLEM
\label{l:gpbotup}
Consider any $\eps > 0$ and any message $\alpha_{i \to j}$ of height $h$. For any $P_{ij}, Q_{ji}, S_{ij}, N_i$ for which $\alpha_{i\to j}\big(P_{ij}, Q_{ji}, S_{ij}, N_i\big)$ is finite and for any $\hat{Q}_{ji}\approxeq_{h'\eps} Q_{ji}$ there exists $\hat{P}_{ij}\approxeq_{h\eps} P_{ij}$ such that
\begin{equation}
\label{e:gpbotup}
\begin{split}
\hat{\alpha}_{i \to j}(\hat{P}_{ij}, \hat{Q}_{ji}, S_{ij}, N_i)~\leq~\exp\big((\max\{h', h\} + h)\eps\big) \alpha_{i \to j}(P_{ij}, Q_{ji}, S_{ij}, N_i).
\end{split}
\end{equation}
\ELEM

\BLEM
\label{l:gptopdown}
Let $\hat{\alpha}_{i \to j}$ be a message of height $h$, and consider any $\hat{P}_{ij}, \hat{Q}_{ji}, \hat{S}_{ij}, \hat{N}_i$ 
for which $\hat{\alpha}_{i\to j}\big(\hat{P}_{ij}, \hat{Q}_{ij},  \hat{S}_{ij}, \hat{N}_i \big)$ is finite. Then the set of observations $S\subseteq V_{ij}$, obtained by solving the approximate dynamic program~\eqref{e:acompose'}-\eqref{e:acompose3} on subtree $T_{ij}$, is such that $S \cap \Delta_{ij} = \hat{S}_{ij}$, $|S| \leq \hat{N}_i$, and for any $Q_{ji}\approx_{h'\eps } \hat{Q}_{ji}$,
\begin{equation}
\begin{split}
\label{e:gptopdown}
R_{ij}\big(Q_{ji}, S\big)~\leq~\exp\big((\max\{h', h\} + h)\eps\big) \hat{\alpha}_{i\to j}(\hat{P}_{ij}, \hat{Q}_{ji}, S_{ij}, N_i),
\end{split}
\end{equation}
and such that the distribution on ${\bf X}_{\Delta_{ij}}$ given observations in $S$ has precision $P_{ij}$ in $T_{ij}$, where $P_{ij}\approx_{h\eps}\hat{P}_{ij}$.
\ELEM

The proofs of Lemmas~\ref{l:gpbotup} and~\ref{l:gptopdown} are almost exactly identical to that of Lemmas~\ref{l:gffbotup} and~\ref{l:gfftopdown} respectively.  The only difference between the error analysis for the element-wise rounding for GFFs and the SVD-based rounding in this section is that while the element-wise rounding error increases by a constant factor due to application of transformation $\mathrm{Marginal}$, the SVD-based rounding error remains unchanged---compare~\eqref{e:margapxgff} to~\eqref{e:margapx}. As a result, the error of the element-wise rounding increases exponentially with the height of the message (Lemmas~\ref{l:gffbotup} and~\ref{l:gfftopdown}) whereas the error of the SVD-based rounding scales linearly. Hence we only provide a brief sketch for the proof of Lemma~\ref{l:gpbotup} (to illustrate the above mentioned difference) and omit the proof of Lemma~\ref{l:gptopdown} altogether.

\BPRF[of Lemma~\ref{l:gpbotup}:]
Proof is by induction on the depth $h$. The base case, $h = 1$, is identical to that in the proof of Lemma~\ref{l:gffbotup}.

Consider a message $\alpha_{i \to j}$ of height $h > 1$. Let $P_{ki}^*, Q_{ik}^*, S_{ik}^*, N_k^*, P_{li}^*,Q_{il}^*, S_{il}^*, N_l^*, L_{ij}^*$ be the optimal choice in~\eqref{e:compose'} for $P_{ij} = \hat{P}_{ij},~Q_{ji} = \hat{Q}_{ji},~S_{ij} = \hat{S}_{ij},~N_i = \hat{N}_i$. Let $\hat{S}_{ik}=S_{ik}^*$, $\hat{S}_{il}=S_{il}^*$, $\hat{N}_k=N_k^*$,  $\hat{N}_l=N_l^*$, and $\hat{L}_{ij} = L_{ij}^*$.

Consider matrices $\hat{P}_{ki}$ and $\hat{P}_{li}$ in~\eqref{e:acompose'} such that 
\begin{equation}
\label{e:gpbotup4}
\hat{P}_{ki} \approx_{(h-1)\eps} P_{ki}^* \quad \textrm{and}  \quad \hat{P}_{li} \approx_{(h-1)\eps} P_{li}^*,
\end{equation} 
and let $\hat{Q}_{ik}, \hat{Q}_{il}$, and $\hat{P}_{ij}$ be given respectively by~\eqref{e:acompose3} and~\eqref{e:acompose1}. 

Now $\hat{Q}_{ik}$ is obtained by first adding matrices $\Lambda_{V_i}, \hat{P_{li}}, \hat{Q}_{ji}$, and then successively applying transformations $\mathrm{Obs},~\mathrm{Marginal}$ and $\mathrm{Round}_{\eps}$. By definition, $\hat{P}_{li} \approx_{(h-1)\eps} P_{li}^*$ and by hypothesis, $\hat{Q}_{ji} \approx_{h'\eps} Q_{ji}$. Hence by~\eqref{e:addapx},
$$\big(\Lambda_{V_i} + \hat{P_{li}} + \hat{Q}_{ji} \big) \approx_{\max\{h', (h-1)\} \eps} \big(\Lambda_{V_i} + P_{li}^* + Q_{ji} \big).$$

Since application of $\mathrm{Obs}$ and $\mathrm{Marginal}$ do not increase the rounding error (see~\eqref{e:obsapx} and~\eqref{e:margapx} respectively), we have
\begin{equation}
\label{e:gpbotup3}
\begin{split}
&\mathrm{Marginal}_{V_i \setminus (\hat{S}_{ij} \cup \hat{L}_{ij}),~\Delta_{ik} \setminus \hat{S}_{ik}}\bigg(\mathrm{Obs}\bigg( \Lambda_{V_i} + \hat{P}_{li}  +\hat{Q}_{ji},~\hat{S}_{ij} \cup \hat{L}_{ij} \bigg) \bigg) \\ 
& \qquad \approx_{\max\{h', (h-1)\}\eps}~\mathrm{Marginal}_{V_i \setminus (\hat{S}_{ij} \cup L_{ij}^*),~\Delta_{ik} \setminus S_{ik}^*} \bigg(\mathrm{Obs}\bigg( \Lambda_{V_i} + P_{li}^* + Q_{ji},~S_{ij} \cup L_{ij}^* \bigg) \bigg) \\
& \qquad =~Q_{ik}^*.
\end{split}
\end{equation}

Applying $\mathrm{Round}_{\eps}$ to the l.h.s. of~\eqref{e:gpbotup3} gives us, by the ``triangle inequality" (i.e., Observation~\ref{o:triangle}) that 
\begin{equation*}
\begin{split}
\hat{Q}_{ik}~&=~\mathrm{Round}_{\eps} \bigg( \mathrm{Marginal}_{V_i \setminus (\hat{S}_{ij} \cup \hat{L}_{ij}),~\Delta_{ik} \setminus \hat{S}_{ik}}\bigg(\mathrm{Obs}\bigg( \Lambda_{V_i} + \hat{P}_{li}  +\hat{Q}_{ji},~\hat{S}_{ij} \cup \hat{L}_{ij} \bigg) \bigg) \bigg) \\
~&\approx_{\max\{(h'+1), h\}\eps}~Q_{ik}^*.
\end{split}
\end{equation*}

By a similar analysis, using Lemma~\ref{l:gpops} and Observation~\ref{o:triangle}, we have
$$\hat{Q}_{il}\approx_{\max\{(h'+1), h\}\eps} Q_{il}^* \quad \textrm{and} \quad \hat{P}_{ij}\approx_{h\eps} P_{ij}^*.$$

To complete the induction step we compare the corresponding terms on the r.h.s. of equations~\eqref{e:acompose'} and~\eqref{e:compose'}. Since $\hat{Q}_{ik} \approx_{\max\{(h'+1), h\}\eps} Q_{ik}^*$, by induction hypothesis there exists $\hat{P}_{ki}$ which satisfies both~\eqref{e:gpbotup4} as well as the following: 
\begin{equation}
\label{e:gpbotup1}
\begin{split}
\frac{\hat{\alpha}_{k \to i}(\hat{P}_{ki}, \hat{Q}_{ik}, S_{ik}^*, N_k^*)}{\alpha_{k \to l}(P_{ki}^*, Q_{ik}^*, S_{ik}^*, N_k^*)}~&\leq~\exp\big(\big(\max\{\max\{(h'+1), h\},(h-1)\} + (h-1)\big)\eps\big) \\
~&\leq~\exp\big(\big(\max\{h', h\} + h\big)\eps\big).
\end{split}
\end{equation}

Similarly since $\hat{Q}_{il} \approx_{\max\{(h'+1), h\}\eps} Q_{il}^*$, by induction hypothesis there exists $\hat{P}_{li}$ which satisfies both~\eqref{e:gpbotup4} and
\begin{equation}
\label{e:gpbotup5}
\frac{\hat{\alpha}_{l \to i}(\hat{P}_{li}, \hat{Q}_{il}, S_{il}^*, N_l^*)}{\alpha_{l \to l}(P_{li}^*, Q_{il}^*, S_{ik}^*, N_k^*)}~\leq~\exp\big(\big(\max\{h', h\} + h\big)\eps\big).
\end{equation}

Only the trace terms in~\eqref{e:acompose'} and~\eqref{e:compose'} remain to be compared. It follows from combining~\eqref{e:gpbotup4} and the hypothesis $\hat{Q}_{ji} \approx_{h'\eps} Q_{ji}$ with Observation~\ref{o:traceapprox} and Lemma~\ref{l:gpops} that the ratio of the trace terms,
\begin{equation}
\label{e:gpbotup2}
\frac{\TR \bigg( \big({\hat{\Lambda}'}_{V_{i}} \big[\hat{V}'_i , \hat{V}'_i \big] \big)^{-1} \bigg)}
{\TR \bigg( \big({\Lambda'}_{V_{i}} \big[ V'_i, V'_i \big] \big)^{-1} \bigg)}~\leq~\exp\big(\max\{h', (h-1)\} \eps\big).
\end{equation}

Combining~\eqref{e:gpbotup1},~\eqref{e:gpbotup5} and~\eqref{e:gpbotup2} we obtain~\eqref{e:gpbotup}. 
\EPRF

Since the error scales linearly in the height of the message, we do not need to run our message passing algorithm on shallow tree-decompositions (unlike GFFs for which we used Bodlaender's construction, see Lemma~\ref{l:shallowbinary}) in order to get a FPTAS---we can use any tree-decomposition. Our main result in this section can be stated as follows. 

\BTHM
\label{t:gpmain}
There is a dynamic programming algorithm which, for any $\kappa$, and any Gaussian MRF on a graph of $n$ vertices with tree-width bounded by $\kappa$ does the following: for any $1 > \eps' > 0$ and budget $b$,  it outputs a set $S_{b, \eps'}$ such that
$$\mathrm{err}\big(S_{b, \eps'}\big)~\leq (1 + \eps')~\min_{|S| \leq b}~\mathrm{err}\big(S\big).$$
The algorithm runs in time 
$$\bigg(\frac{n \kappa}{\eps'} \frac{\lambda_{max}(\Lambda)}{\lambda_{min}(\Lambda)} \bigg)^{O(\kappa^2)}b^2~+~n 2^{O(\kappa^3)}.$$ 
\ETHM

Note that the running time scales as polynomial in the condition number of the input precision matrix $\Lambda$ for bounded tree-width graphs. This means we obtain an FPTAS  if the condition number is bounded, e.g., by a polynomial in the size (i.e. number of bits) of the description of $\Lambda$---this however is not generally the case.

\BPRF[of Theorem~\ref{t:gpmain}:]
Consider a tree-decomposition $T$ having width $\kappa' = \kappa$ and $m \leq 2n+1$ clusters, and which satisfies our requirements in Note~\ref{note:decompassume} (see page~\pageref{note:decompassume}). We can construct such a $T$ in time $n 2^{O(\kappa^3)}$ by using the algorithm of Bodlaender (see Theorem~1.1 of~\citet{Bod96}).

 Let $\eps = \frac{\eps'}{4(2n + 1)}$. We run the algorithm given by~\eqref{e:acompose'}-\eqref{e:acompose3} and~\eqref{e:agpleaf} on tree $T$, with the transformation $\mathrm{Round}_{\eps}$ and the $\eps$-nets $\big\{{\cal I}_{\eps}^{V \times V} \big\}$ as defined respectively in Lemma~\eqref{l:epsmessage} and in Definition~\ref{def:gpepsnet}. Recall that the cluster $V_m$ as per our assumption is empty, i.e., $V_m = \emptyset$, and is a leaf with neighbour (say) $V_i$ so that $\Delta_{im} = V_i$. Our output, $S_{b, \eps'}$, is simply the set of observations extracted from the approximate message $\hat{\alpha}_{i \to m}\big(\hat{P}_{im}^*, \hat{Q}_{mi}, \hat{S}_{im}^*, b\big)$ where $\hat{Q}_{mi}$ is an all-zeros matrix and where
$$\hat{P}_{im}^*,  \hat{S}_{im}^*~\define~\arg \underset{\hat{P}_{im},~\hat{S}_{im} \subseteq \Delta_{im}}{\min}~\hat{\alpha}_{i \to m}\big(\hat{P}_{im}, \hat{Q}_{mi}, \hat{S}_{im}, b\big).$$

Let the optimal solution be $S_b^*$ and let $P_{im}^*$ be the precision matrix of ${\bf X}_{\Delta_{im}} = {\bf X}_{V_i}$ given observations $S_b^*$. 
Now given our choice of $\eps = \frac{\eps'}{4(2n+1)}$ and given that height of $T$ is at most $2n$, we have
\begin{equation*}
\begin{split}
\mathrm{err}\big(S_{b, \eps'}\big)~&=~\frac{1}{n} R_{im}\big(\hat{Q}_{mi}, S_{b, \eps'}\big) \quad (\textrm{by the definition of}~R_{im},~\textrm{see~\eqref{e:approxerr})} \\
~&\leq~\exp(\eps'/4)~\frac{1}{n}\hat{\alpha}_{i \to m}\big(\hat{P}_{im}^*, \hat{Q}_{mi}, \hat{S}_{im}^*, b\big) \quad (\textrm{by Lemma~\ref{l:gptopdown}}~\textrm{and our choice of}~\eps) \\
~&\leq~\exp(\eps'/4)~\frac{1}{n}\hat{\alpha}_{i \to m}\big(\hat{P}_{im}, \hat{Q}_{mi}, S_b^* \cap \Delta_{im}, b\big) \quad (\textrm{where}~\hat{P}_{im}~\textrm{is as defined}  \\
~& \qquad \qquad \qquad \qquad \textrm{in Lemma~\ref{l:gpbotup} for arguments}~P_{im}^*, \hat{Q}_{mi}, S_b^* \cap \Delta_{im}, b) \\
~&\leq~\exp(\eps'/4)\frac{1}{n} \bigg( \exp(\eps'/4)~\alpha_{i \to m}\big(P_{im}^*, \hat{Q}_{mi}, S_b^* \cap \Delta_{im}, b\big)\bigg) \\
~& \qquad \qquad \qquad \qquad \qquad \qquad \qquad \qquad (\textrm{by Lemma~\ref{l:gpbotup}}~\textrm{and our choice of}~\eps) \\
~&<~(1 + \eps')~\frac{1}{n}\alpha_{i \to m}\big(P_{im}^*, \hat{Q}_{mi}, S_b^* \cap \Delta_{im}, b\big)\bigg) \\
~&=~(1 + \eps')\mathrm{err}\big(S_b^*\big).
\end{split}
\end{equation*}

As for the running time, first recall that the time required to construct $T$ is $n 2^{O(\kappa^3)}$. Next, note that all precision matrices produced by message passing have supports of size at most $\kappa \times \kappa \leq \kappa^2$, and hence the size of each $\eps$-net used (as per~\eqref{e:sizeeps}) is at most
$$O\bigg( \bigg(\frac{n^2 \kappa^3}{\eps^2}\bigg(\frac{\lambda_{max}(\Lambda)}{\lambda_{min}(\Lambda)} \bigg)^2 \bigg)^{\kappa^2} \bigg),$$
which, given $\eps = \frac{\eps'}{4(2n+1)}$, is of the order $O\bigg( \bigg(\frac{n^4 \kappa^3}{\eps'^2}\bigg(\frac{\lambda_{max}(\Lambda)}{\lambda_{min}(\Lambda)} \bigg)^2 \bigg)^{\kappa^2} \bigg)$. Hence using sparse representations for matrices, we can perform the message passing step~\eqref{e:acompose'} for each edge $\{i, j\}$ in $T$ in time 
$$\bigg(\frac{n \kappa}{\eps'} \frac{\lambda_{max}(\Lambda)}{\lambda_{min}(\Lambda)}  \bigg)^{O(\kappa^2)}~b^2,$$

giving the claimed time complexity since by construction $T$ has at most $2n$ edges.

\EPRF

\bibliographystyle{chicago}
\bibliography{gpjournal}


\end{document}

%% file: definitions.tex
\def\eps{\varepsilon}

\def\ee{\mathbf{E}}
\def\var{\mathbf{V}}

\def\prec{\Lambda} 
\def\define{\overset{\mathrm{def}}{=}}

\def\TR{\mathrm{Tr}\,}

\def\R{{\mathbb R}}

\def\e{{\mathrm e}}

\def\<{\langle}
\def\>{\rangle}
\def\({\left(} 
\def\){\right)} 

\newtheorem{thm}{Theorem}
\newtheorem{que}[thm]{Question}
\newtheorem{pro}[thm]{Proposition}
\newtheorem{obs}[thm]{Observation}
\newtheorem{exc}{Exercise}
\newtheorem{nte}{Note}
\newtheorem{rem}[thm]{Remark}
\newtheorem{lem}[thm]{Lemma}
\newtheorem{cor}[thm]{Corollary}

\newtheorem{con}[thm]{Conjecture}
\newtheorem{exm}{Example}
\newtheorem{dfn}[thm]{Definition}
\newtheorem{fact}[thm]{Fact}

\newenvironment{prf}[1]{\noindent{\bf{Proof #1\\}}}{$\hfill\blacksquare$\nopagebreak[4]\vskip 0.3cm}

\def\BS{\begin{footnotesize}}
\def\ES{\end{footnotesize}}
\def\BCON{\begin{con}}
\def\ECON{\end{con}}
\def\BOBS{\begin{obs}}
\def\EOBS{\end{obs}}

\def\BPRO{\begin{pro}}
\def\EPRO{\end{pro}}
\def\BIDE{\begin{ide}}
\def\EIDE{\end{ide}}
\def\BTHM{\begin{thm}}
\def\ETHM{\end{thm}}
\def\BDEF{\begin{dfn}\rm}
\def\EDEF{\end{dfn}}
\newcommand\BPRF[1][:]{\begin{prf}{#1}}
\def\EPRF{\end{prf}}
\def\BLEM{\begin{lem}}
\def\ELEM{\end{lem}}
\def\BEX{\begin{exm}\rm}
\def\EEX{\end{exm}}
\def\BEXC{\begin{exc}\rm}
\def\EEXC{\end{exc}}
\def\BCOR{\begin{cor}}
\def\ECOR{\end{cor}}
\def\BQUE{\begin{que}}
\def\EQUE{\end{que}}
\newcommand\BSOL[1][:]{\begin{sol}{#1}}
\def\ESOL{\end{sol}}
\def\BNTE{\begin{nte}}
\def\ENTE{\end{nte}}
\def\BFAC{\begin{fact}}
\def\EFAC{\end{fact}}

\def\BIT{\begin{itemize}}
\def\EIT{\end{itemize}}
\def\BREM{\begin{rem}\rm}
\def\EREM{\end{rem}}
\def\BC{\begin{center}}
\def\EC{\end{center}}
\def\BEQ{\begin{equation}}
\def\EEQ{\end{equation}}
